\documentclass[acmtog,authorversion,nonacm]{acmart}
\usepackage[utf8]{inputenc}
\usepackage[table]{xcolor} 
\usepackage{graphicx}      
\usepackage{multirow}      

\usepackage[ruled]{algorithm2e} 

\SetAlFnt{\small}
\SetAlCapFnt{\small}
\SetAlCapNameFnt{\small}
\SetAlCapHSkip{0pt}

\usepackage{amsmath,amsthm}

\usepackage{graphicx}
\usepackage{textcomp}
\usepackage{subcaption}
\usepackage{xspace}
\usepackage{enumitem}
\usepackage{hyperref}
\usepackage{cleveref}
\usepackage{multirow}
\usepackage{dblfloatfix} 

\definecolor{blue}{rgb}{0,0,1}
\definecolor{red}{rgb}{1,0,0}
\definecolor{green}{rgb}{0,.5,0}
\definecolor{darkgreen}{rgb}{0,.4,0}
\definecolor{orange}{rgb}{0.75, 0.4, 0}
\definecolor{teal}{rgb}{0.0, 0.4, 0.4}
\definecolor{purple}{rgb}{0.65,0,0.65}
\definecolor{black}{rgb}{0,0,0}

\usepackage[ruled]{algorithm2e} 

\SetAlFnt{\small}
\SetAlCapFnt{\small}
\SetAlCapNameFnt{\small}
\SetAlCapHSkip{0pt}

\begin{document}

\title{Progressive Photorealistic Simplification}

\author{Adi Rosenthal}
\orcid{0000-0000-0000-0000}
\affiliation{%
 \institution{Reichman University and Google}
\country{Israel}
}
\email{adirosenthal@google.com}

\author{Dana Berman}
\orcid{0000-0002-1265-3864}
\affiliation{%
 \institution{Google}
\country{Israel}
}
\email{danaberman@google.com}

\author{Yedid Hoshen}
\orcid{0000-0002-0967-4541}
\affiliation{%
 \institution{Hebrew University and Google}
\country{Israel}
}
\email{yedid.hoshen@mail.huji.ac.il}

\author{Ariel Shamir}
\orcid{0000-0001-7082-7845}
\affiliation{%
 \institution{Reichman University and Google}
\country{Israel}
}
\email{arik@runi.ac.il}

\begin{abstract}
Existing image simplification techniques often rely on Non-Photorealistic Rendering (NPR), transforming photographs into stylized sketches, cartoons, or paintings. While effective at reducing visual complexity, such approaches typically sacrifice photographic realism. In this work, we explore a complementary direction: simplifying images while preserving their photorealistic appearance.
We introduce progressive semantic image simplification, a framework that iteratively reduces scene complexity by removing and inpainting elements in a controlled manner. At each step, the resulting image remains a plausible natural photograph. Our method combines semantic understanding with generative editing, leveraging Vision-Language Models (VLMs) to identify and prioritize elements for removal, and a learned verifier to ensure photorealism and coherence throughout the process. This is implemented via an iterative \emph{Select–Remove–Verify} pipeline that produces high-quality simplification trajectories.
To improve efficiency, we further distill this process into an image-to-video generation model that directly predicts coherent simplification sequences from a single input image. Beyond generating cleaner and more focused compositions, our approach enables applications such as content-aware decluttering, semantic layer decomposition, and interactive editing.
More broadly, our work suggests that simplification through structured content removal can serve as a practical mechanism for guiding visual interpretation within the photorealistic domain, complementing traditional abstraction methods.
\end{abstract}

\ccsdesc[500]{Computing methodologies~Computational photography}
\ccsdesc[500]{Computing methodologies~Scene understanding}
\ccsdesc[500]{Computing methodologies~Image manipulation}
\ccsdesc[500]{Computing methodologies~Non-photorealistic rendering}


\begin{teaserfigure}
\centering
\includegraphics[width=1\textwidth]{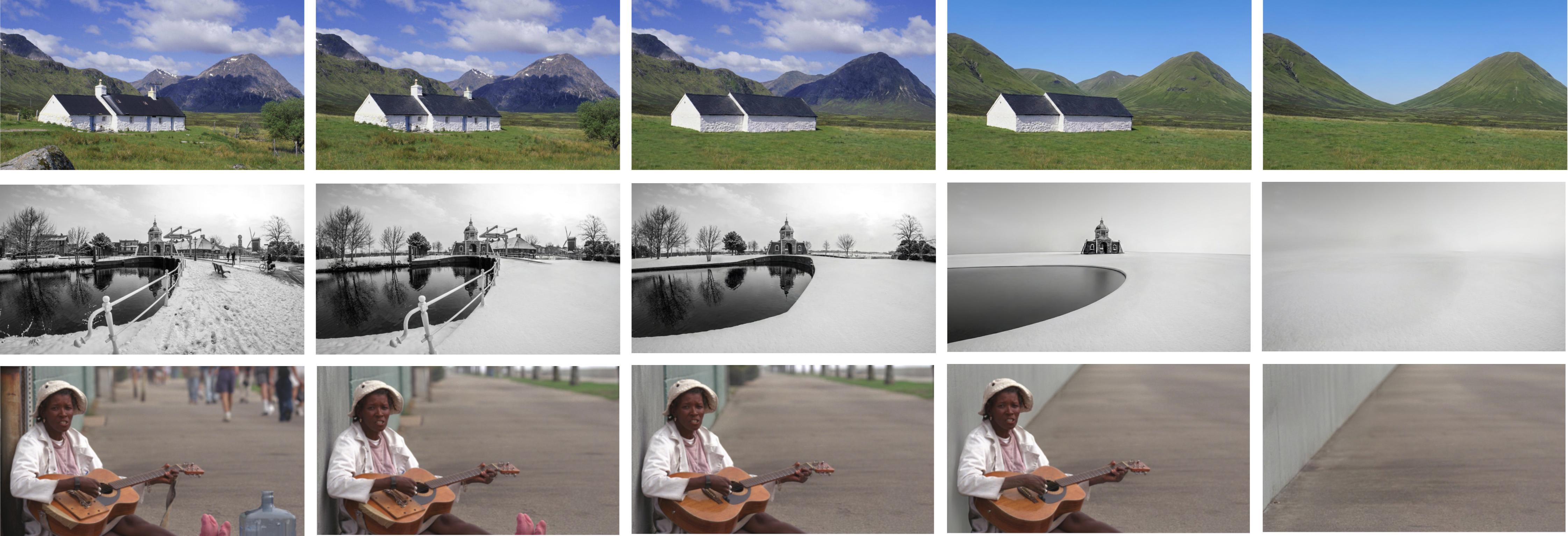}
\caption{
Our method progressively abstracts a photograph by intelligently removing scene elements. Starting with a complex original image (left), our pipeline iteratively generates a sequence of simplified images. Crucially, each image in the sequence remains photorealistic, resulting in more minimalist compositions (towards the right) that focus on the essence of the original scene, or even go further to totally abstract images.
}
\label{fig:teaser}
\end{teaserfigure}


\sloppy
\maketitle
\section{Introduction}
\label{sec:intro}

Photographic abstraction and depiction have long been central themes in art, vision science, and computer graphics. Beyond their computational treatment in recent decades, the notions of realism, stylization, and abstraction have been studied for centuries in artistic practice and theory. From Renaissance investigations of perspective, shading, and composition, to later analyses in art history and perception~\cite{gombrich1960art,kemp1990science,Livingstone2002vision,Marr2010vision,Palmer1999vision}, abstraction has been understood as a process of distilling visual scenes, emphasizing salient structure while suppressing extraneous detail.

A key insight from this body of work is that visual perception is inherently interpretive. As Gombrich argued, ``there is no innocent eye'': understanding an image depends on prior knowledge, expectations, and perceptual organization. Similarly, 
Gestalt theory further highlights how humans organize visual input into meaningful structures, prioritizing certain elements over others~\cite{wertheimer1938laws,Palmer1999vision}.
From this perspective, abstraction is not merely a stylistic transformation, but a process that guides perception by selectively simplifying and structuring visual information.

In computer graphics, abstraction has been most prominently explored through non-photorealistic rendering (NPR), where images are transformed into stylized representations such as sketches, cartoons, or painterly depictions~\cite{decarlo2002stylization,XDOG-2012,Kang2009FlowBased,Winnemoeller:2006:RTV,hertzmann1998painterly,Lu:2010:IPS,vinker2022clipasso,winkenbach1994computer,curtis1997computer,10.1145/383259.383295}. These methods achieve simplification primarily through changes in visual style, often departing from photorealism.

In this work, we explore a complementary perspective: we concentrate on \emph{simplification} while still preserving photorealism. 
Rather than altering style, we consider abstraction as the progressive removal of elements that are less relevant to the scene’s interpretation, while preserving both realism and semantic coherence.

We formalize this problem as \emph{semantic image simplification}, a process that reduces scene complexity by removing or suppressing elements in a controlled manner. In our approach, simplification operates directly on the scene content rather than through stylistic reinterpretation.
We introduce a framework for progressive subtractive simplification in which scene elements, ranging from foreground objects to background details, are removed step by step. The central challenge is to determine \emph{what} to remove and \emph{when}, such that each intermediate result remains a plausible photograph and continues to convey the intended visual content.

To address this, we propose a two-stage approach. In the first stage, we construct simplification trajectories using a search-based pipeline that iteratively applies a \emph{Select-Remove-Verify} loop. A vision-language model identifies candidate elements for removal, a generative editing model performs the modification, and a learned verifier evaluates whether the resulting image remains photorealistic and semantically consistent. This stage produces high-quality but computationally expensive sequences.

In the second stage, we distill these trajectories into a generative model by fine-tuning an image-to-video diffusion transformer to predict the simplification sequence from a single input image. This enables efficient inference while maintaining coherence across steps.

Our approach allows users to interactively explore alternative simplification paths by selecting different removal targets during the process. Beyond improving visual clarity and composition, this framework supports applications such as content-aware decluttering and semantic layer decomposition. More broadly, our work connects classical ideas of abstraction and perception with modern generative modeling, showing how progressive simplification can serve as a computational mechanism for guiding visual interpretation while preserving photorealism.

\begin{figure}[t]
    \centering
    \includegraphics[width=\linewidth]{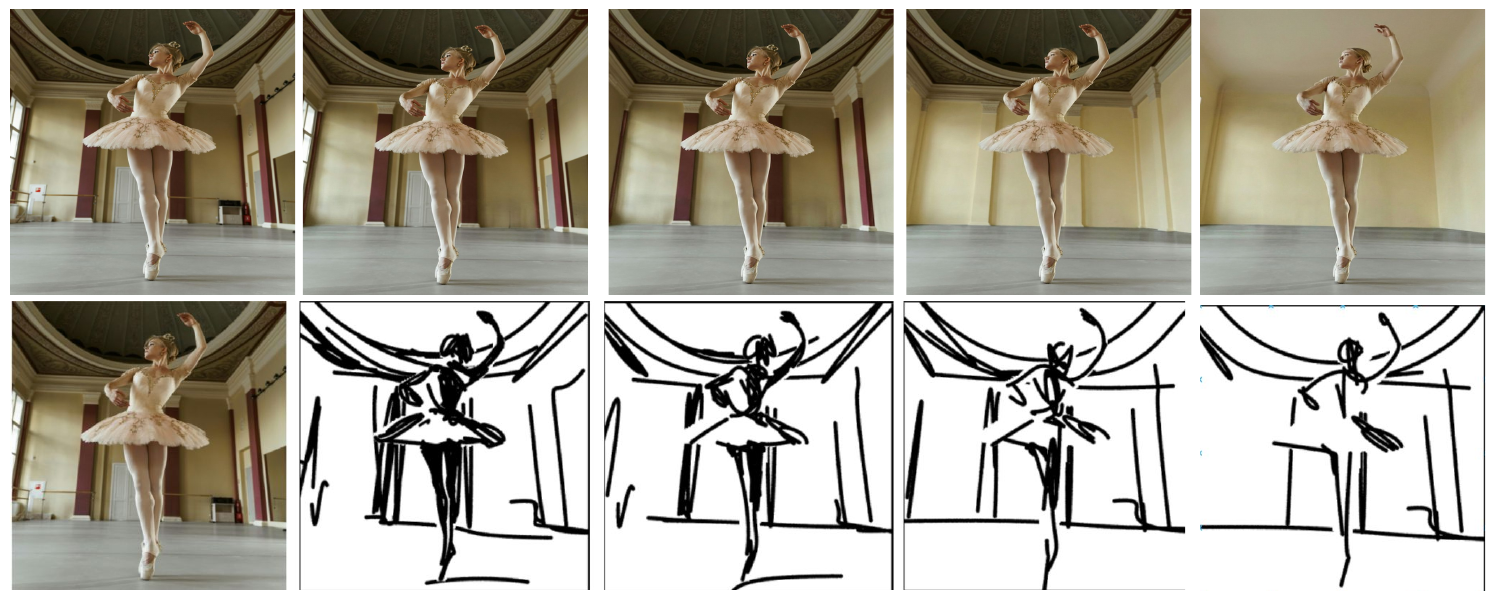}
    \caption*{Comparison with progressive sketch-based Simplification.}
    \label{fig:compare_sketch}
    
    \vspace{0.1cm} 
    
        \includegraphics[width=\linewidth]{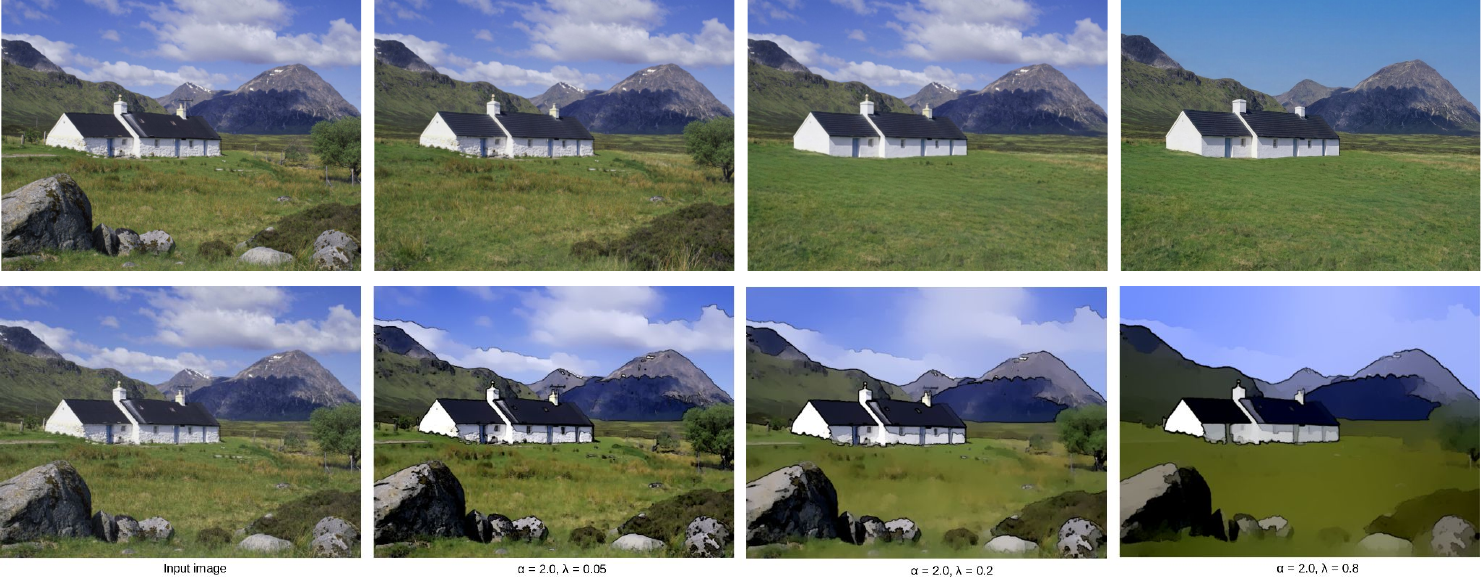}
        \caption*{Comparison with progressive region Simplification.}
        \label{fig:compare_style}
    
    \caption{\textbf{Photorealistic vs. Stylistic Simplification.} 
    We compare our method (top rows) against two progressive simplification methods (bottom rows): CLIPascene sketch abstraction~\cite{vinker2023clipascene} and stylistic simplification~\cite{farbman2008edge}. The previous approaches simplify scenes by altering the medium (e.g., to sketches or cartoons), our method simplifies semantic content but preserves photographic fidelity.}
    \label{fig:abstraction_comparison}
\end{figure}

\section{Related Work}
\label{sec:related}

\paragraph{Depiction, Perception, and Abstraction}
The notions of abstraction, stylization, and depiction have long been studied in art theory and vision science. Foundational work in perception emphasizes that visual understanding is not passive, but involves structured organization and interpretation~\cite{gombrich1960art,Marr2010vision,Livingstone2002vision,Palmer1999vision}. In particular, Gestalt principles describe how humans group and prioritize visual elements to form coherent structures~\cite{wertheimer1938laws,Palmer1999vision}. From this perspective, abstraction can be viewed as a process of simplifying visual input to emphasize salient structure and guide perception. These classical concepts were later formalized into foundational principles of computational depiction~\cite{10.1145/508530.508550,willats2005defining}. Our work draws on this broader view, focusing on simplification through controlled removal of scene elements.

\paragraph{Semantic and Stylistic Abstraction}
Computational image abstraction has been predominantly explored in the context of Non-Photorealistic Rendering (NPR), where simplification is achieved through stylization. Early work includes stroke-based rendering~\cite{winkenbach1994computer,hertzmann1998painterly} and data-driven image analogies~\cite{10.1145/383259.383295}, followed by content-aware abstraction using perceptual cues~\cite{decarlo2002stylization}, and real-time cartoon filters~\cite{winnemoller2006real, chen2007real}. Methods such as edge-preserving decomposition~\cite{farbman2008edge} enabled progressive abstraction, while neural approaches like Neural Style Transfer~\cite{gatys2016image} separated content and style. More recent methods, including \textit{CLIPasso}~\cite{vinker2022clipasso}, \textit{CLIPascene}~\cite{vinker2023clipascene}, and \textit{DiffSketcher}~\cite{xing2023diffsketcher}, leverage vision-language and diffusion models to generate sketch-based or stylized representations. However, these approaches fundamentally alter the visual modality (e.g., sketches or paintings). In contrast, our work focuses on simplification \emph{within} the photorealistic domain.

\paragraph{Photorealistic Removal and Decluttering}
Preserving realism during content removal requires accurate segmentation and high-quality inpainting. Classical approaches relied on patch-based synthesis~\cite{criminisi2004region}, while recent foundation models such as SAM~\cite{kirillov2023segment} and LaMa~\cite{suvorov2022resolution} enable large-scale, high-quality editing. Supervised editing methods~\cite{winter2024objectdrop} further improve realism using counterfactual supervision, and recent vision-language models~\cite{nano-banana} support instruction-driven editing. Parallel work on image decluttering has explored identifying and removing distracting elements: \citet{fried2015finding} characterized visual distractors using human annotations, \citet{aberman2022deep} introduced deep saliency priors, and \textit{SimpSON}~\cite{huynh2023simpson} automated distractor detection. Our work builds on these advances, but differs in its \emph{progressive} and \emph{structure-aware} simplification of scenes.

\paragraph{Progressive Simplification and I2V Priors}
A small number of recent works explore sequential object removal. For example, ``Visual Jenga''~\cite{bhattad2024visual} removes objects based on physical stability, focusing on scene dynamics. In contrast, we prioritize \emph{semantic importance}, aiming to preserve the interpretability of the scene throughout the simplification process. Our approach further leverages Image-to-Video (I2V) foundation models (e.g.,~\cite{wan2025}), which encode strong physical and temporal priors. Recent works~\cite{wiedemer2025video, wu2025chronoedit, zhang2025image} have used such models for general-purpose editing; we instead employ them to distill a carefully constructed, but computationally expensive, simplification process into an efficient generative model.

\section{Problem Formulation}
\label{sec:problem_formulation}

We define \textit{photorealistic subtractive simplification} as the process of sequentially removing scene elements to reveal the essential structure of an image. Our mathematical formulation models this problem as photorealistic trajectory generation.

Given an initial image $I_0$ of a real scene $\mathcal S$ composed of $n$ elements $\mathcal{E} = \{e_1..e_n\}$, which can be objects or background elements, our goal is to determine a removal permutation $\pi$ ordered by increasing semantic importance. By sequentially removing elements according to $\pi$ we generate a trajectory of photorealistic images $I_0,..,I_n$ where each step represents a progressively deeper simplification of the scene. Formally, this can be modeled as finding an optimal path from the root to a leaf in a conceptual tree $\mathcal{T}$ , where the root is $I_0$ and each edge in the tree represents the removal of a single element. Generally, a node at level $k$ has $n-k$ children and each path from the root to a leaf represents one of the $n!$ possible element permutations. 

Finding the optimal path in  $\mathcal{T}$ presents two cardinal challenges:
\begin{enumerate}
\item \textbf{Computational Complexity:} The search space ($n!$ permutations) is super-polynomial, making exhaustive enumeration impossible.
\item \textbf{Subjectivity of Ordering:} There is no robust analytical scoring function to rank trajectories. The relative importance of objects is often ambiguous. For example, a shoe and a sock in a cluttered room may be equally ``distracting,'' with no strict objective ordering between them.
\end{enumerate}

\subsection{A Simple Model for Subtractive Simplification}
\label{sec:taxonomy}

\begin{figure}[t]
    \centering

    \includegraphics[width=1.0\linewidth]{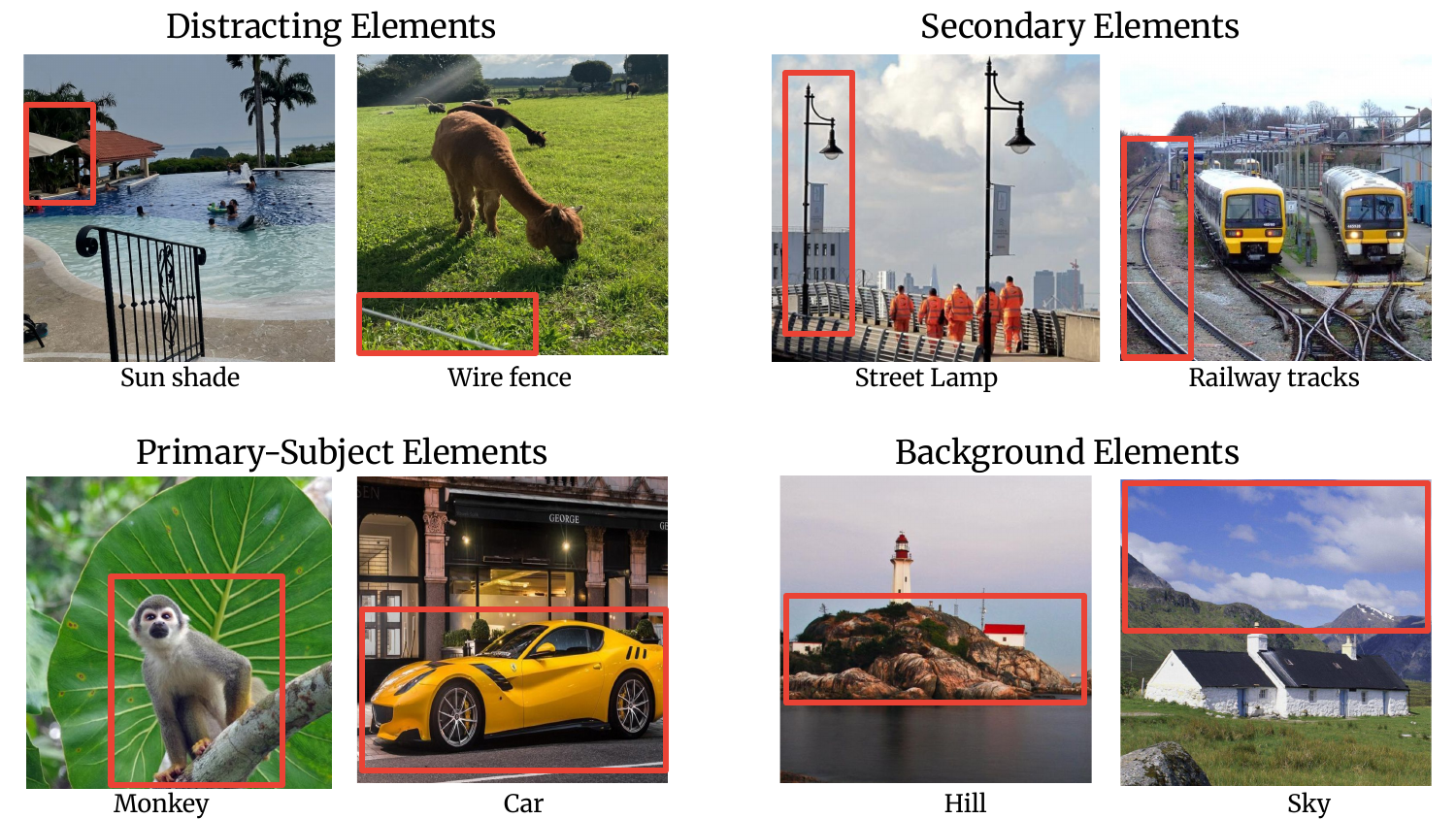}

    \caption{\textbf{Semantic classification of image elements.} The figure depicts elements from each class, whose description is written below the image and marked by a bounding box.}
    
    \label{fig:element_classes}
\end{figure}

Ideally, the removal order $\pi$ should be determined by fine-grained human preference. However, since strict object-to-object ordering is often subjective, we instead propose a simple, interpretable schema (taxonomy) that correlates with human preference. While this schema provides a coarse ranking, it offers a robust ``scaffold'' for the simplification process. We heuristically group the objects in $\mathcal{E}$ into four semantic levels of importance (see examples in Fig.~\ref{fig:element_classes}): 
\begin{enumerate}
\item \textbf{Distracting Elements:} Objects that actively interfere with the aesthetic or composition (e.g., trash, stray cables). These are removed first.
\item \textbf{Secondary Elements:} Contextual objects that support the scene but are not essential (e.g., a vase on a table).
\item \textbf{Primary-Subject Elements:} The core subjects; removing them fundamentally alters the scene's semantic identity.
\item \textbf{Background Elements:} The environment itself, serving as the final layer of simplification (removed last).
\end{enumerate}

\paragraph{Taxonomy Validation.} To verify alignment with human perception, three independent raters annotated 93 images. Inter-rater agreement was strong (Kendall's $\tau_b \in [0.63, 0.75]$) ~\cite{kendall1938new}. Crucially, the semantic levels proved distinct: confusion between hierarchy extremes (``Primary'' vs. ``Background'') was negligible ($<2.5\%$), as shown in Fig.~\ref{fig:combined_analysis} (see supplementary for matrix construction details).

We further validated the removal order via pairwise comparisons. When comparing elements across \textit{different} categories, consensus was overwhelming (Table~\ref{tab:element_removal_hybrid}): ``Distractors'' were removed first in 100\% of cases, and ``Secondary'' elements were prioritized over ``Primary'' subjects in 96.1\% of cases. Conversely, comparisons within the \textit{same} category showed high ambiguity, with raters finding the order irrelevant in 27.5\% of cases. This validates our hybrid design: the taxonomy serves as a robust high-level scaffold, while fine-grained ranking requires the semantic reasoning of a VLM. We provide further experimental details and analysis in the SM.

\begin{figure}[t]
  \centering
  \includegraphics[width=1.0\linewidth]{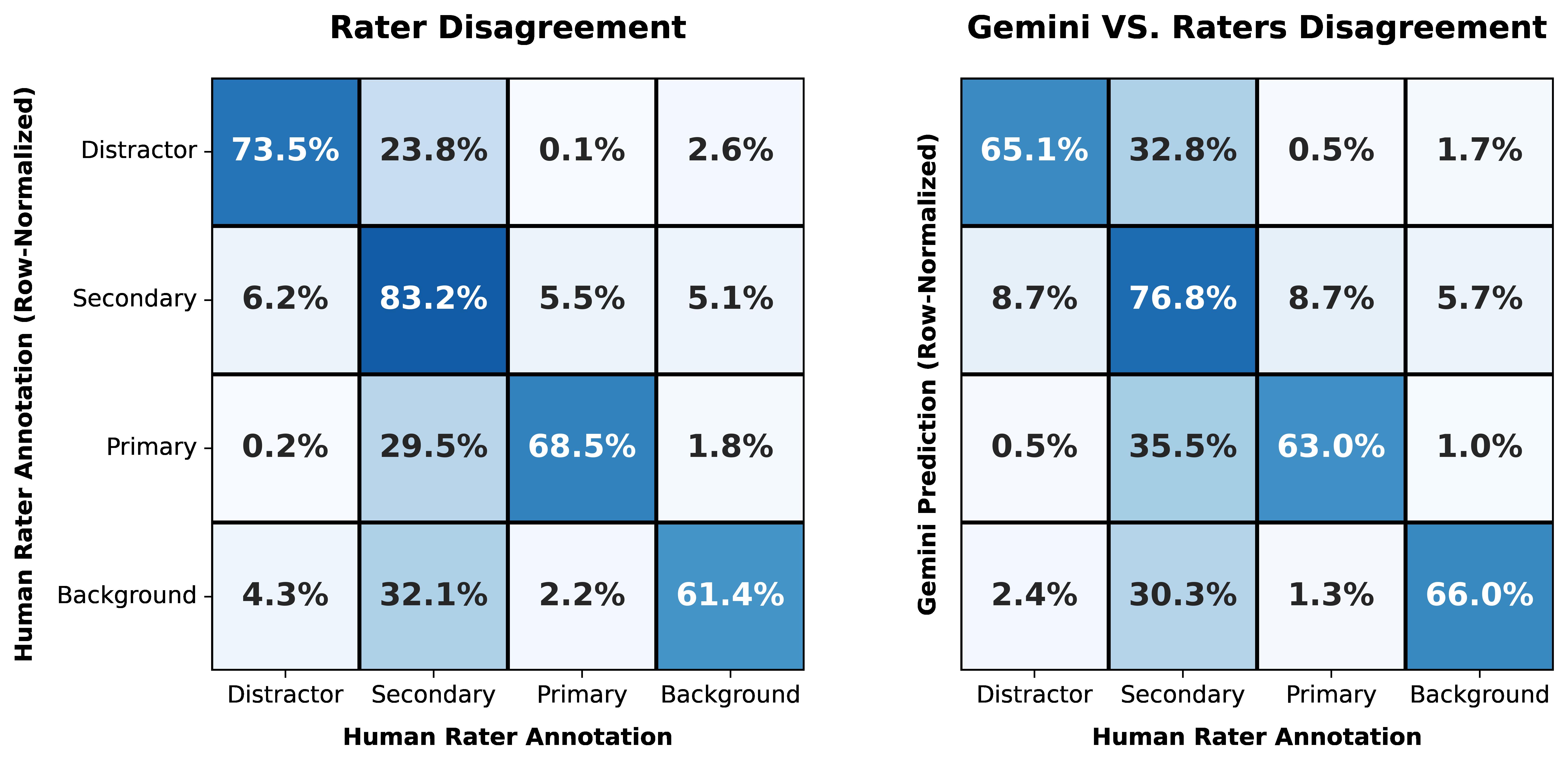}
\caption{\textbf{Comparison of Classification Agreement.} (Left) Row-Normalized Aggregated Confusion Matrix of Human Raters (Inter-rater reliability) demonstrates strong consensus on the ``Secondary'' class (83.2\%), though raters frequently conflate ``Background'' elements with ``Secondary'' (32.1\%). (Right) Disagreement Analysis of Gemini vs. Human Raters. While Gemini generally aligns with human perception, it exhibits a stronger bias toward the ``Secondary'' class, incorrectly assigning it to ``Primary'' (35.5\%) and ``Distractor'' (32.8\%) objects more frequently than human inter-rater baselines.}
  \label{fig:combined_analysis}
\end{figure}

\begin{table}[h]
    \centering
    \caption{\textbf{Pairwise removal preference} (Row vs. Column). In our perceptual study, raters were asked to choose which element they would remove first. The values represent the normalized frequency that the Row element was selected over the Column element. For example, Secondary elements were prioritized for removal over Primary subjects in 96.1\% of cases. This strong directional consensus validates the ordering of our proposed hierarchy.}
    \label{tab:element_removal_hybrid}
    
    \setlength{\tabcolsep}{5pt}
        \begin{tabular}{lcccc}
        \toprule
        & \textbf{Dist.} & \textbf{Sec.} & \textbf{Prim.} & \textbf{Back.} \\
        \midrule
        \textbf{Distractor} & --     & 100.0\% & 100.0\% & 100.0\% \\
        \textbf{Secondary}  & 0.0\%  & --      & 96.1\%  & 87.5\%  \\
        \textbf{Primary}    & 0.0\%  & 3.9\%   & --      & 81.5\%  \\
        \textbf{Background} & 0.0\%  & 12.5\%  & 18.5\%  & --      \\
        \bottomrule
    \end{tabular}
\end{table}

\begin{figure*}[t] 
    \centering 
    \includegraphics[width=\linewidth]{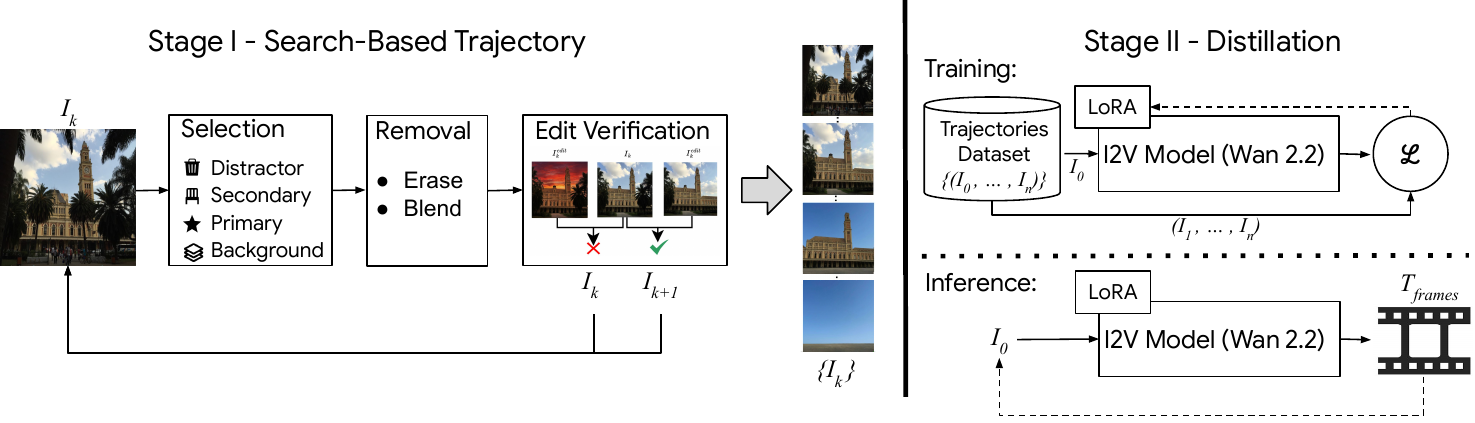} 
    \caption{ \textbf{Method Overview.} Our framework operates in two stages to achieve effective progressive photorealistic simplification. \textbf{Stage I (Search-Based Trajectory)} is an iterative, closed-loop pipeline. A VLM-based \textit{Planner} identifies scene elements for removal based on semantic rank. The \textit{Robust Execution} module performs inpainting (Generate, Align, Blend), while a \textit{Classifier Edit Verification} step ensures the edit preserves photorealism before accepting the result ($I_{k+1}$). \textbf{Stage II (Distillation)} distills these verified trajectories into a feed-forward video generation model. We fine-tune a pre-trained I2V model (Wan 2.2) using LoRA on the dataset of valid trajectories. For visualization purposes, we depict the training objective in pixel space, though the actual fine-tuning is performed in the latent diffusion objective. During \textit{Inference}, the model predicts subtractive stop-motion videos from a single input image ($I_0$).} 
    \label{fig:architecture} 
\end{figure*}

\section{Method}
\label{sec:method}

We propose a two-stage approach. First, we develop a search-based algorithm to generate high-quality simplification trajectories (Sec.~\ref{sec:search_method}). While effective, this process is computationally intensive. Therefore, in the second stage, we distill these verified trajectories into a fast video generation model for efficient inference (Sec.~\ref{sec:distillation}).

\subsection{Stage I: Search-Based Trajectory Generation}
\label{sec:search_method}

We propose an iterative \textit{Select-Remove-Verify} algorithm. This method combines the coarse guidance of our taxonomy with the fine-grained reasoning of VLMs to determine the precise removal order.

\paragraph{1. Element Selection.} We use a Vision–Language Model (VLM) as a high-level planner. At iteration $k$, the VLM (Gemini 3 Pro) analyzes the current image $I_k$, constrained by the current active category from our taxonomy (starting with \textit{Distractor}). While the taxonomy provides the broad category, we require higher resolution to select the specific next object. The VLM evaluates all candidates within the active level and identifies the single element with the least semantic importance. By re-planning at every step rather than adhering to a fixed list from $I_0$, the system adapts to changes in composition and occlusions revealed by previous edits.

\paragraph{2. Element Removal.} Removing an object while preserving the background requires high-precision inpainting. We use a universal editing model (Gemini 3 Pro Image) to generate candidate edits given the chosen element description.  However, generative editors often induce global color shifts or introduce subtle artifacts that accumulate over successive steps, leading to a drift in overall image quality (examples are in the supplementary material). To mitigate this, we employ a robust blending pipeline to ensure the edits are well localized:
\begin{enumerate}
\item \textbf{Alignment:} We register the edited candidate to $I_k$ using keypoint matching and warping.
\item \textbf{Masking:} We compute a difference mask, weighted by the inverse local gradient to suppress changes near strong edges (preserving structural integrity).
\item \textbf{Blending:} We perform local histogram matching and soft alpha blending within the mask, copying all other pixels directly from $I_k$.
\end{enumerate}

\paragraph{3. Verification.} To filter element removal failures, we employ a learned classifier to gate edits. At each step, the system generates up to five candidates, accepting the first that passes the classifier; otherwise, the removal is skipped. The model is trained on a balanced dataset of $\sim$8,000 pairs, comprising curated positives and diverse negatives (manual pipeline failures, reversed trajectories, and synthetic adversarial edits).

We utilize a Siamese DINOv2-Large backbone adapted with LoRA ($r=16$). To explicitly model local inconsistencies, extracted patch tokens are reshaped into feature grids $G \in \mathbb{R}^{H \times W \times D}$ to compute a difference map $\Delta = |G_{in} - G_{out}|$. The concatenated volume $[G_{in}; G_{out}; \Delta]$ is processed by a light-weight 2-layer CNN head. 

\subsection{Stage II: Distillation via Video Generation}
\label{sec:distillation}

The iterative search process described above yields high-quality data but is too slow for real-time applications. To address this, we distill the logic of the search-based planner into a faster image-to-video generation model. We treat the sequence of abstracted images $(I_0, \dots, I_n)$ as a subtractive ``stop-motion'' video, where the temporal dimension represents simplification depth.

\paragraph{Architecture and Training.}We leverage \emph{Wan 2.2}, a 14B parameter Image-to-Video (I2V) latent diffusion model. We freeze the base model and use Low-Rank Adaptation (LoRA) \cite{hu2022lora} modules to specific attention and feed-forward layers. We train the model on the validated trajectories generated in Stage I. Long trajectories are uniformly sub-sampled to a fixed length $T_{\text{train}}$. The model is trained with a standard diffusion objective to predict noise $\epsilon$, conditioned on the initial frame $I_0$. This effectively teaches the model to internalize the semantic taxonomy and removal logic of the search algorithm.

\paragraph{Iterative Inference.}At test time, the model takes a single input $I_0$ and generates a clip of $T_{\text{test}}$ frames. Because complex scenes require more steps than a single clip allows, we apply the model autoregressively: the final frame of the generated clip is fed back as the input for the next segment. This results in a continuous video where early frames perform conservative decluttering and later frames converge toward minimal compositions, retaining the high quality of the search method at a fraction of the computational cost.

\section{Experiments and Results}
\label{sec:experiments}

\vspace{1em}

\subsection{Experimental Setup}
\label{sec:setup}

\paragraph{Training data}. We used high-quality, diverse, in-the-wild images from CADB~\cite{zhang2021image} as trajectory starting points. For the \textbf{classifier} in Stage I, we gathered overall 8000 training images, some manually curated and some automatically generated via synthesis and process reversal (adding elements to images). For Stage II \textbf{distillation}, we generated sequences with variable lengths using the Search-based Trajectory Generation of Stage I. After manually reviewing the trajectories and discarding low-quality samples, 500 trajectories remained. Since the length of each sequence depends on the scene complexity, we sampled subsequences of 10 removal steps, and generated standard-length videos of 49 frames, by repeating each frame 5 times (with the final step truncated to 4 frames). The resulting dataset consists of 4,500 high quality subtractive stop motion sequences.

\paragraph{Implementation details}
\label{sec:implementation}
The verification classifier model used LoRA fine-tuning of DINOv2-Large network for 16 epochs on a single NVIDIA A100 GPU using the AdamW optimizer ($lr=2 \times 10^{-4}$). For inference in Stage I we use a V100 GPU.
For distillation, we fine-tuned Wan 2.2 (14B). We used separate LoRA adapters ($r=32, \alpha=32$) on all attention and MLP layers, and on both experts. Training used AdamW ($lr=1\times 10^{-4}$) on 4$\times$ NVIDIA A100 (80GB) GPUs with a batch size of 1, taking 4-7 days per expert. Inference used 10 diffusion steps and a single GPU.   

\begin{figure*}[t] 
  \centering
  \setlength{\tabcolsep}{2pt} 
  
  \begin{tabular}{ c c } 
     
    \textbf{Ours - Search-Based.} & \textbf{Ours - Distilled.} \\[1mm]

    \includegraphics[width=0.49\linewidth,height=0.12\linewidth]{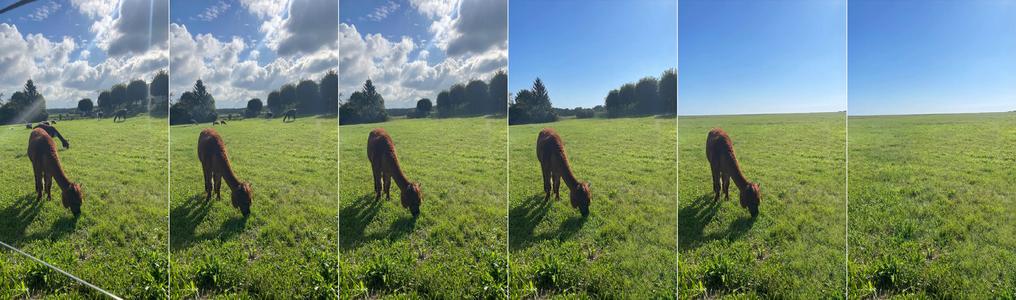} &
    \includegraphics[width=0.505\linewidth,height=0.12\linewidth]{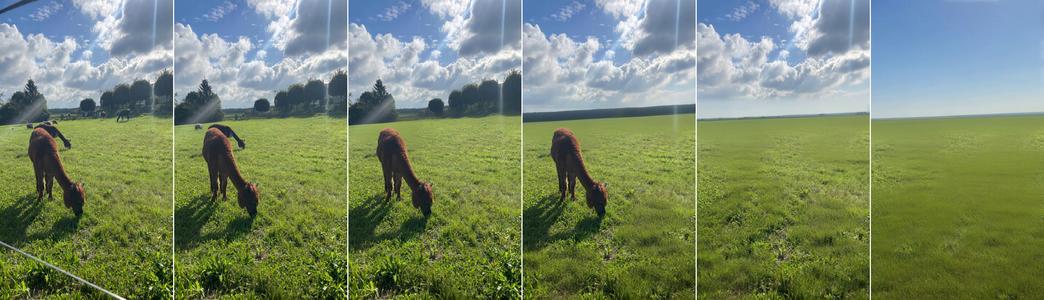} \\

    \includegraphics[width=0.49\linewidth,height=0.12\linewidth]{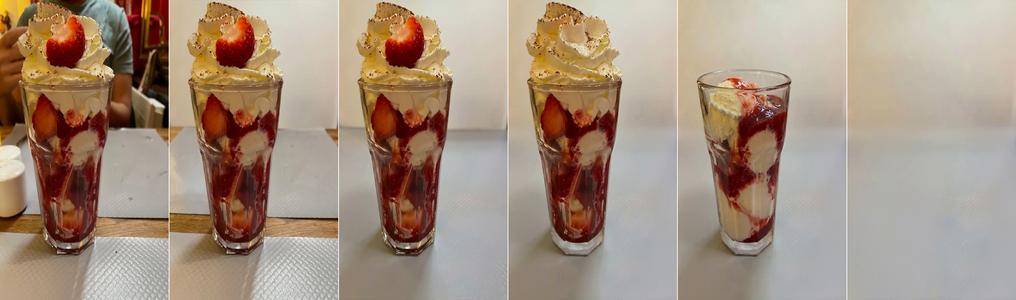} &
    \includegraphics[width=0.505\linewidth,height=0.12\linewidth]{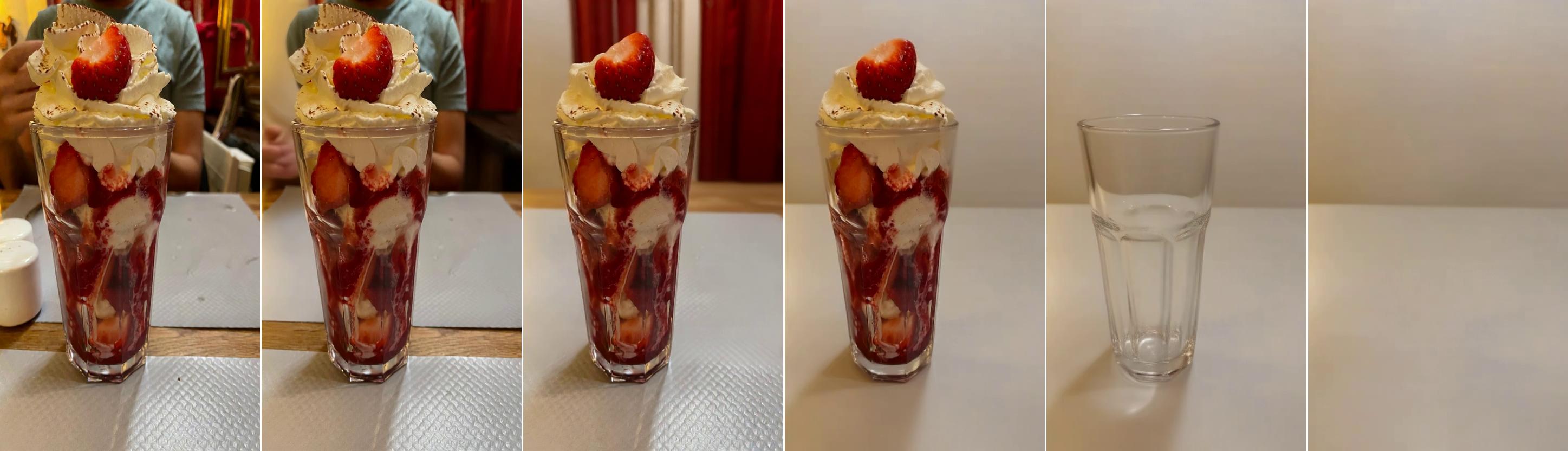} \\
     
    \includegraphics[width=0.49\linewidth,height=0.12\linewidth]{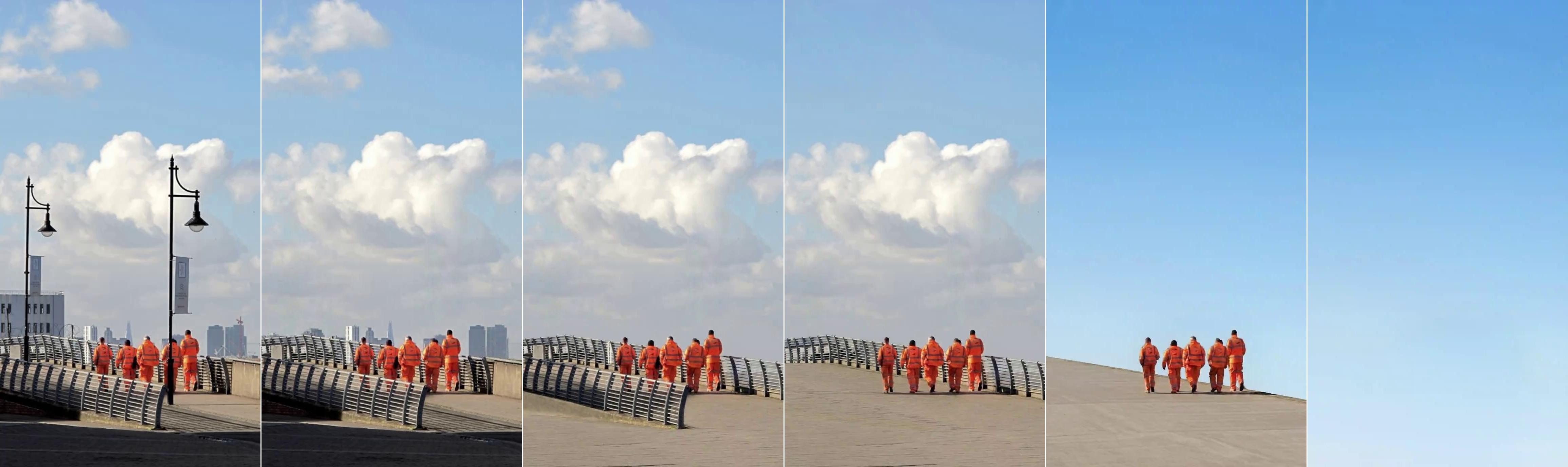} &
    \includegraphics[width=0.505\linewidth,height=0.12\linewidth]{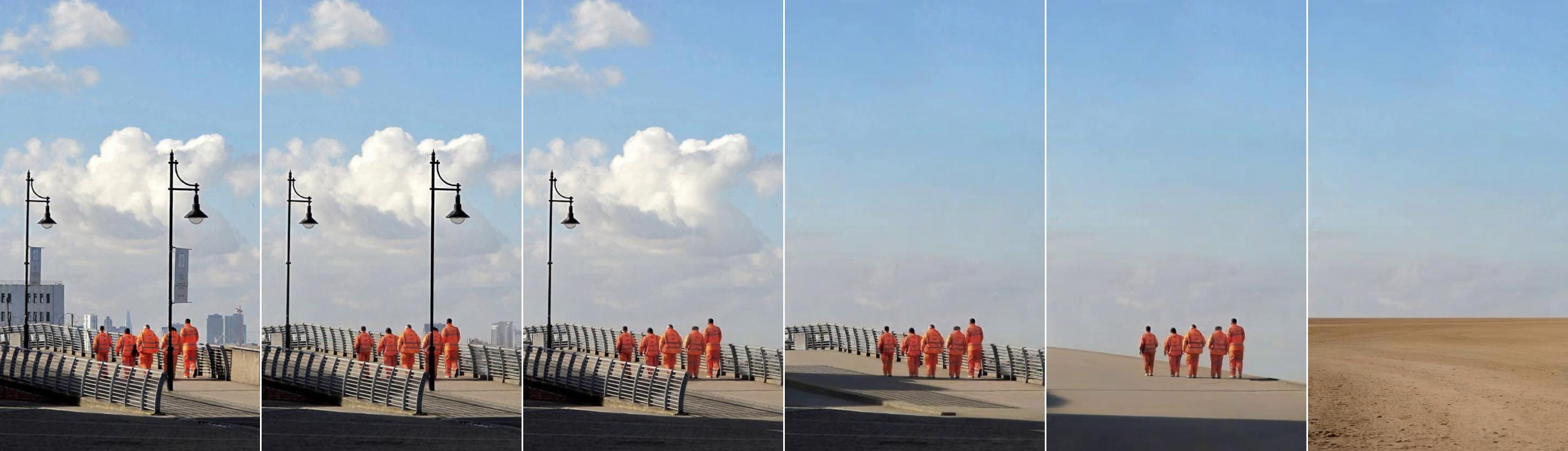} \\

  \end{tabular}
  \caption{\textbf{Qualitative Comparison.} Visual results comparing our Search-Based method (Left) against the Distilled approach (Right). The distilled model maintains temporal consistency while successfully and iteratively removing scene elements.}
  \label{fig:comare_stages}
  
\end{figure*}

\subsection{Evaluation Methodology}
\label{sec:eval_methodology}

\subsubsection*{Evaluation datasets and ground truth}\label{par:ground_truth}

Our test set is a subset of $93$ images from Open Images~\cite{openimages}. Three independent human raters classified the scene elements per our taxonomy (Sec.~\ref{sec:taxonomy}). To increase reliability, the final ground truth for each image is derived from the \textit{strict intersection} of annotations from the two most aligned raters (as determined by inter-rater agreement analysis). This process yields a set of $530$ validated objects $\mathcal{O}_{GT}$ in 68 images (average of 7.8 objects per image), each assigned a strict semantic rank. We also collected precise, human-verified segmentation masks for every object in the ground truth set.

To evaluate subtractive simplification, we determine if objects are eliminated in the semantic order defined by our rigorous ground truth. This requires a two-stage automated pipeline: first, detecting \textit{when} a specific object is removed in the generated video, and second, scoring that removal order of all objects against the ground truth (GT) hierarchy.

We developed an automated pipeline to identify the specific frame $t^*$ where an object $O$ is removed.
We first compute a binary difference mask for each time step. We then verify the object's removal if the mask sufficiently covers the object area and a significant change occurs \textit{inside} the object region, while the background remains stable. Please refer to Section C in the supplemental file for more details.

\subsubsection*{Evaluation Metric: Pairwise Order Accuracy}
\label{sec:metrics}

To quantify the alignment between the detected removal sequence and the ground-truth hierarchy, we use a modified version of Kendall’s $\tau_b$~\cite{kendall1948rank,kendall1938new}, which is better suited for our discrete class ordinal taxonomy.

Let $N'$ be the number of object pairs whose GT class is different. As explained in the taxonomy definition in Sec~\ref{sec:taxonomy}, comparing objects within the same class is meaningless. Each object is defined by $(t*, c)$, where $t*$ is the timestamp in which it was removed, and $c$ is the GT class. We compute the following quantities for different error types: 

\begin{itemize}[leftmargin=2em, noitemsep]
    \item $N_{\text{inv}}$: Counts the occurrences where a high-importance object is removed before a low-importance one ($t_{\text{high}} < t_{\text{low}}$).
    \item $N_{\text{eq}}$: Counts the  occurrences where the model fails to distinguish layers, resulting in simultaneous removal of objects with different importance ($t_{\text{high}} = t_{\text{low}}$).
\end{itemize}

The final \textit{Order Accuracy} score represents the ratio of correctly ordered pairs:
\begin{equation*}
  \text{Order Accuracy} = 1 - \frac{N_{\text{inv}} + N_{\text{eq}}}{N'}.
\end{equation*}

\subsection{Results}

\paragraph{Quantitative comparison.} Tab.~\ref{tab:results} compares our method to $2$ off-the-shelf leading image-to-video generation models: Veo 3.1 (I2V) and Wan 2.2 (I2V). To ensure a fair evaluation, all models were tested using the exact same text prompt.
Pretrained video generation models struggle with the subtractive simplification task. 
Wan 2.2 achieves a low Order Accuracy of $0.214$, indicating near-random ordering of element removal. 
Veo 3.1 demonstrates moderate improvement with a score of 0.402, yet still frequently fails to adhere to the strict semantic hierarchy (e.g., removing background elements before the main subject). 
In contrast, our proposed method demonstrates substantial alignment with the ground-truth simplification logic, achieving a high score of $0.728$.

\begin{table}[t]
\centering
\caption{\textbf{Quantitative Comparison of Semantic Ordering.} We report the \textbf{Order Accuracy}, measuring the strict adherence to the correct object removal hierarchy.} 

\label{tab:results}
\begin{tabular}{l c}
\toprule
\textbf{Method} & \textbf{Order Accuracy $\uparrow$} \\
\midrule
Wan 2.2 (Base) & 0.214 \\
Veo 3.1 & 0.402 \\
\textbf{Ours} & \textbf{0.728} \\
\bottomrule
\end{tabular}

\end{table}

\paragraph{Qualitative results} We demonstrate the efficacy of our approach across diverse scene categories in Figs.~\ref{fig:qualitative_part1} and \ref{fig:qualitative_part2} as well as in the supplementary file. These results validate that our model adheres to a consistent semantic hierarchy: clutter and transient objects are removed first, followed by secondary elements, preserving the primary subject until the final stages. The inpainting maintains high photorealism and structural consistency. Please refer to our \textbf{Supplementary Video} to observe the full subtractive simplification effect of the generated sequences, which exhibit stable transitions between simplification levels.

\paragraph{Distillation efficiency} While Stage I generates high-resolution data, it is computationally expensive as it makes many VLM and inpainting calls. To quantify the improvement of distillation, we ran inference using both methods on a test set of 106 complex and diverse images.  Stage I averages over two hours per image, extending beyond four hours for intricate scenes. In contrast, the distilled model (Stage II) with 2 iterative inference steps requires an average of 15 minutes on the same data. 

\subsection{Ablation}

\paragraph{Trajectory filtering ablation} We analyzed 3 variants of our distilled model. (i) Unfiltered - finetuning Wan2.2 on all the trajectories ($\sim$7,500 sequences for 15 epochs) without filtering. (ii) SFT - initializing with the final weights of the Unfiltered run and finetuning on $\sim$1,500 manually selected, high-quality trajectories. This is inspired by the standard practice in LLM training. (iii) HQ-Only - training the model just on the manually selected trajectories for 46 epochs, ensuring our model never sees any bad trajectories. The results are in Tab.~\ref{tab:ablations}.

\paragraph{Verification classifier ablation.} The classifier acts as a strict visual gate to prevent artifacts and preserve visual quality. Table 3 evaluates its impact on Order Accuracy, which measures \textit{semantic ordering}, not visual fidelity. Ablating the classifier confirms this filter does not degrade semantic ordering; rather, visual artifacts can disrupt the scene and may result in a lower score. See the supplementary material for examples of negative predictions.

\begin{table}[h]
\centering
\caption{\textbf{Ablation results (order accuracy).} We ablate the effects of removing trajectory filtering, and removing the verification classifier. Results are reported across 5 experimental rounds (seeds / temperature)}.
\label{tab:ablations}
\begin{tabular}{l c}
\toprule
\textbf{Method} & \textbf{Order Accuracy $\uparrow$} \\
\midrule
Distilled - Unfiltered & 0.641 $\pm 0.012$ \\
Distilled - SFT & 0.684 $\pm 0.026$ \\
Distilled - HQ-Only & 0.716 $\pm 0.021$\\
Search-based w/o classifier & 0.723 $\pm 0.027$ \\ 
Search-based &  \textbf{0.728} $\pm 0.019$ \\
\bottomrule
\end{tabular}
\end{table}


\section{Applications}
\label{sec:applications}

Our approach enables novel workflows in computational photography and creative design. We highlight three primary applications.

\begin{figure}[t] 
    \centering
    \includegraphics[width=\linewidth]{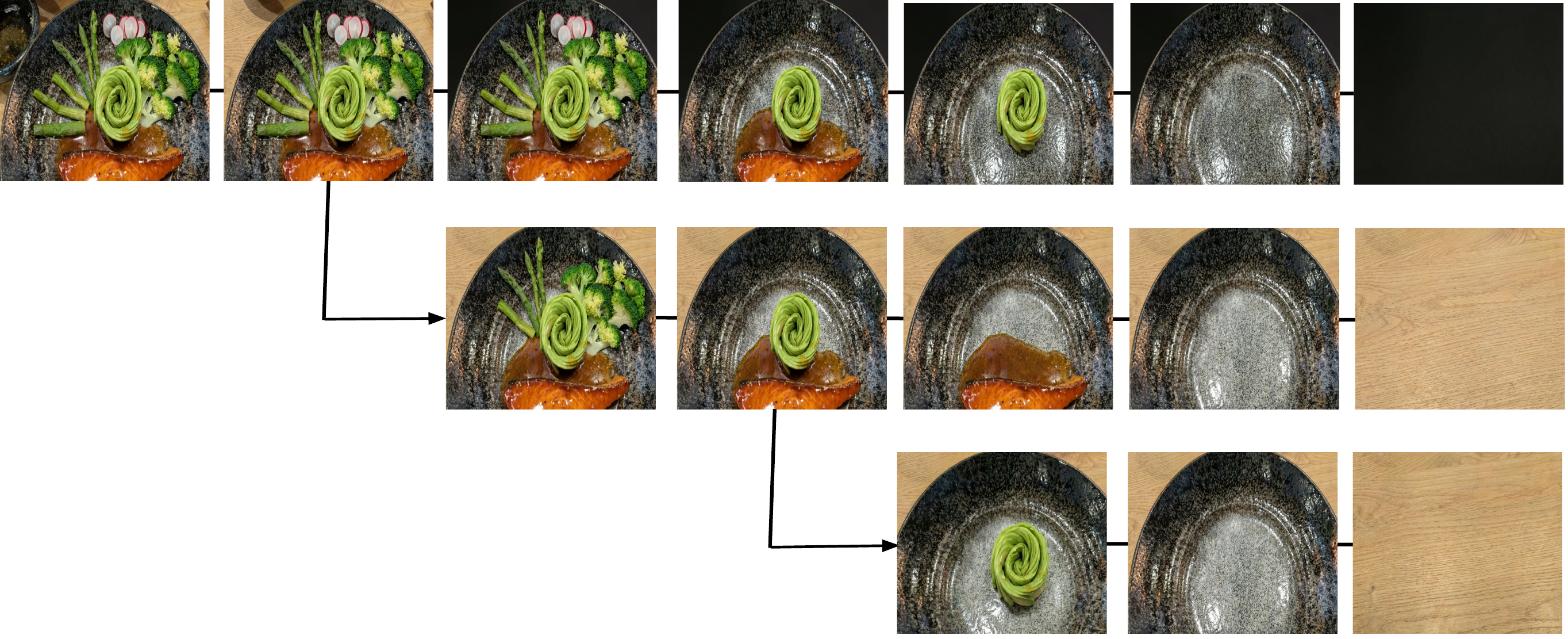}
\caption{\textbf{Interactive Simplification Trees.} 
    Visualizing the edit "tree" from Sec.~\ref{sec:applications}. 
    Our interactive planner allows the user to intervene at intermediate states. 
    Here, the user chooses to ``branch'' the sequence to preserve the wooden table texture, and then removing one object over the other, demonstrating the system's flexibility in defining subjective visual hierarchies.}
    \label{fig:tree_edits}
\end{figure}
\begin{figure}
    \centering

\label{fig:decluttering}
    \includegraphics[width=\linewidth,height=0.8\linewidth]{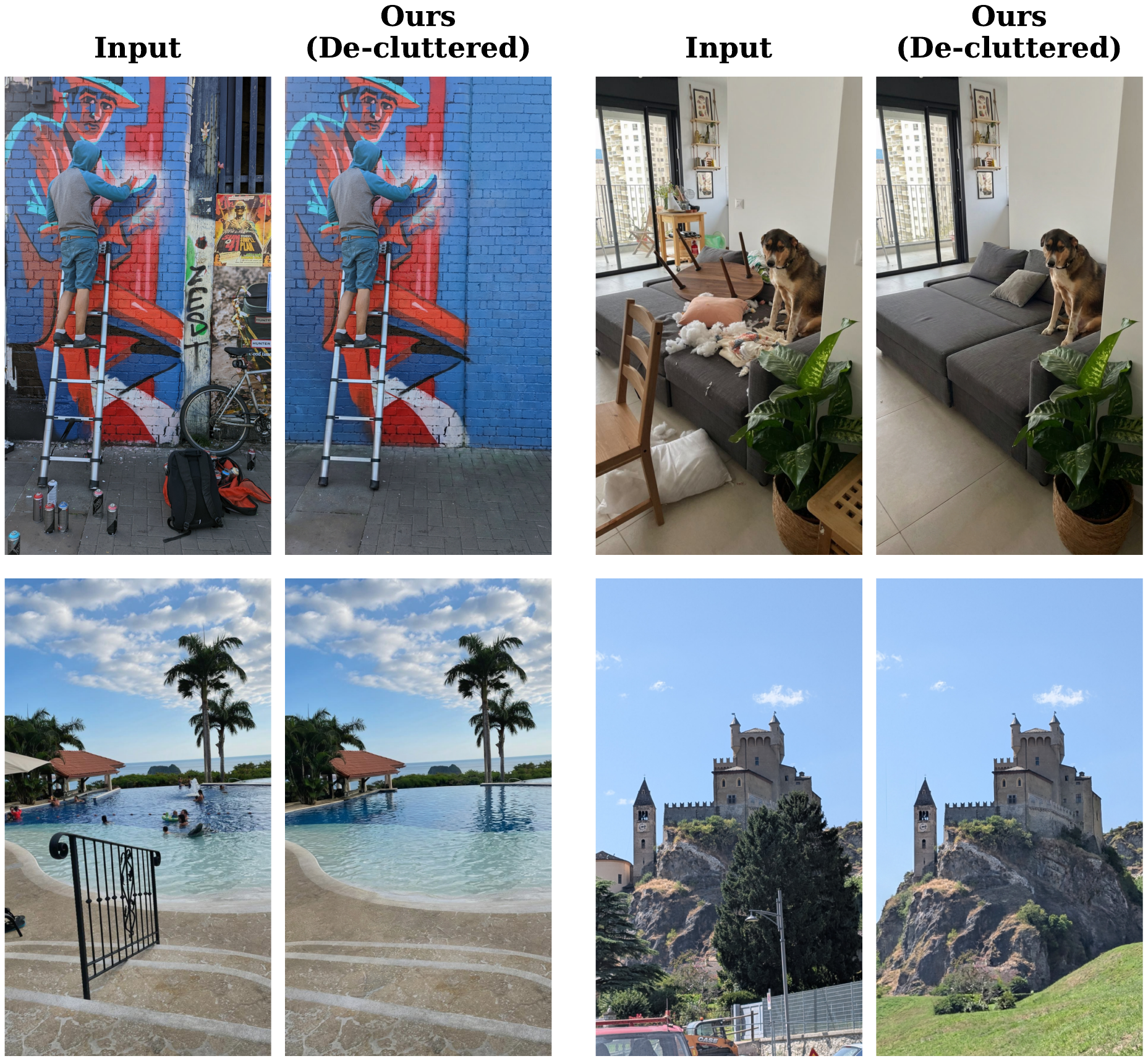}   
      \caption{\textbf{Semantic Image De-cluttering.} Our approach automatically identifies and removes visual noise, such as power lines, transient crowds, and background litter, without requiring manual masks or user prompts. The model produces clean, aesthetic compositions while strictly preserving the primary subject and scene context.}
    \label{fig:diclutter}
\end{figure}

\paragraph{Progressive photorealistic simplification.} Our framework's core application is exploring a scene's visual hierarchy, effectively transforming a static photograph into a dynamic simplification ``slider.'' Because visual importance is subjective, our system empowers users to define a scene's ``essence'' through interactive trajectory editing. Using our Stage I search-based planner, in contrast to the fast, feed-forward inference of the distilled model, users can intervene at any intermediate state to retain or remove specific objects, overriding the automated proposal. By ``branching'' the simplification tree at these decision points and regenerating the subsequent sequence, users can explore alternative aesthetics and tailor the process to their specific narrative or artistic vision, as shown in Fig.~\ref{fig:tree_edits}.

\begin{figure*}[t] 
 \centering
  \includegraphics[width=\linewidth]{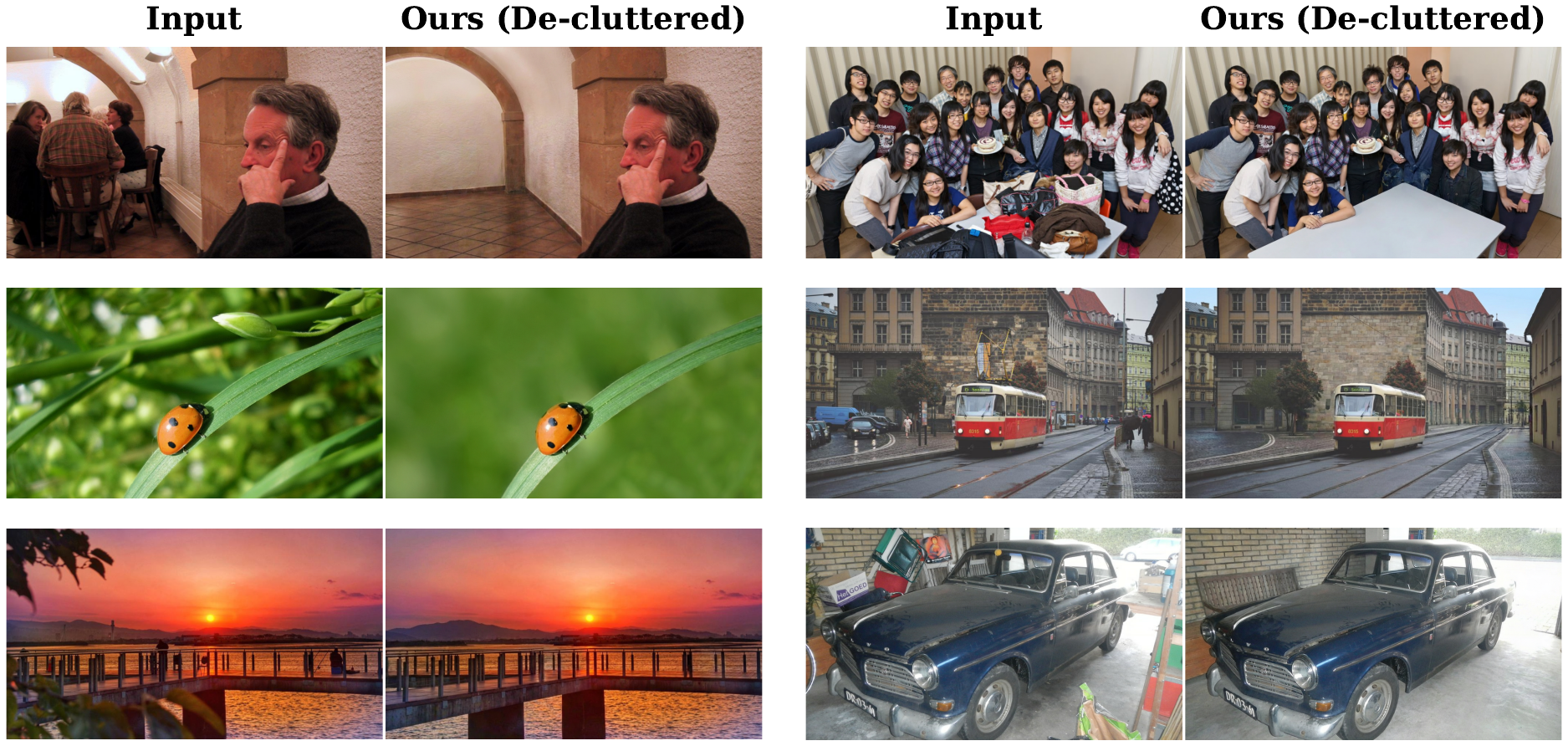}
  \caption{\textbf{Semantic Image De-cluttering.} Our approach automatically identifies and removes visual noise without requiring manual masks or user prompts. The model produces clean, aesthetically enhanced compositions while strictly preserving the primary subject and scene context.}
  \label{fig:diclutter2} 
\end{figure*}

\paragraph{Semantic image de-cluttering}
A common challenge in photography is the presence of unwanted visual noise, wires, litter, transient objects, or cluttered background elements, that detract from the main subject. Manual removal of these elements via inpainting is often tedious and requires technical expertise. Our model offers an automated solution for content-aware de-cluttering, see~\Cref{fig:diclutter,fig:diclutter2}. Because our semantic taxonomy prioritizes the removal of ``Distracting'' and ``Secondary'' elements in the earliest stages of generation, the initial frames of our output video automatically serve as cleaned-up versions of the input. Unlike generic object removal tools that require manual masking, our approach leverages its internalized semantic understanding to identify and remove clutter proactively, enhancing aesthetic quality while preserving the core subject and environment.

\begin{figure}[t]
  \centering
  \includegraphics[width=0.8\linewidth]{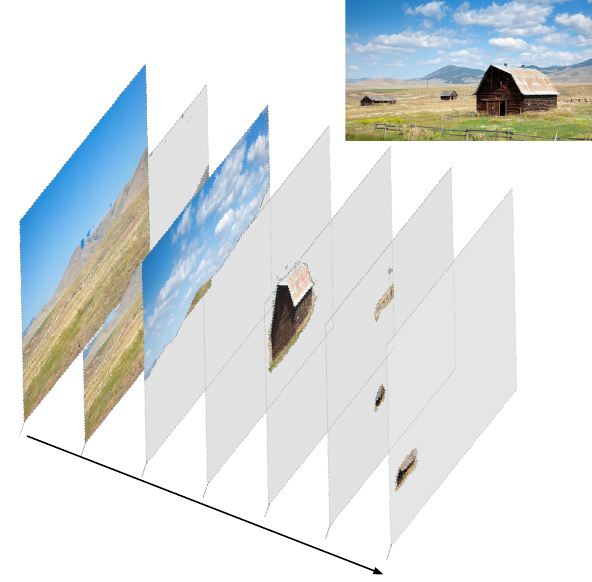}
  \caption{\textbf{Automatic Layer Extraction.} Our method automatically decomposes the input image into independent semantic layers. Based on visual importance, we isolate the scene, enabling object-level editing and scene reconstruction.}

  \label{fig:exploded_view}
\end{figure}

\paragraph{Image layering and semantic decomposition}
Compositing and visual effects workflows often require decomposing a flat image into constituent semantic layers, e.g., background only, background plus secondary objects, background plus primary subjects. Our progressive removal process performs this decomposition automatically. By stepping backwards through the generated removal sequence, we can isolate different semantic strata of the scene, see \Cref{fig:exploded_view}. The final frame represents the pure background layer; preceding frames add primary subjects back into the scene, followed by secondary context. This sequential output provides artists with pre-inpainted semantic layers, significantly streamlining tasks such as parallax creation for 2.5D photo animations, re-lighting specific scene elements, or inserting new objects behind existing ones.

\section{Conclusion}
\label{sec:conclusion}

We proposed a new approach for image simplification by progressively removing elements from the scene while keeping the results photorealistic and presented a method to create such progressive simplifications from input images. 

Our method has its limitations. While the distillation model remains relatively slow, we anticipate that the emergence of faster foundation models will significantly reduce video inference time. The order of selection for removal can still be debatable, but our interactive tree exploration can alleviate that. Moreover, the Stage I removal and inpainting processes rely heavily on the base model, which may occasionally fail or introduce visual artifacts. 

A more general message of our work is that with today’s powerful models, abstraction does not have to be non-photorealistic. Instead, computational tools now enable new directions in visual simplification, creative editing, and exploratory photography, where realism and abstraction coexist along a continuous spectrum. 



\begin{figure*}[t]
  \centering
  
    \includegraphics[width=0.95\linewidth]{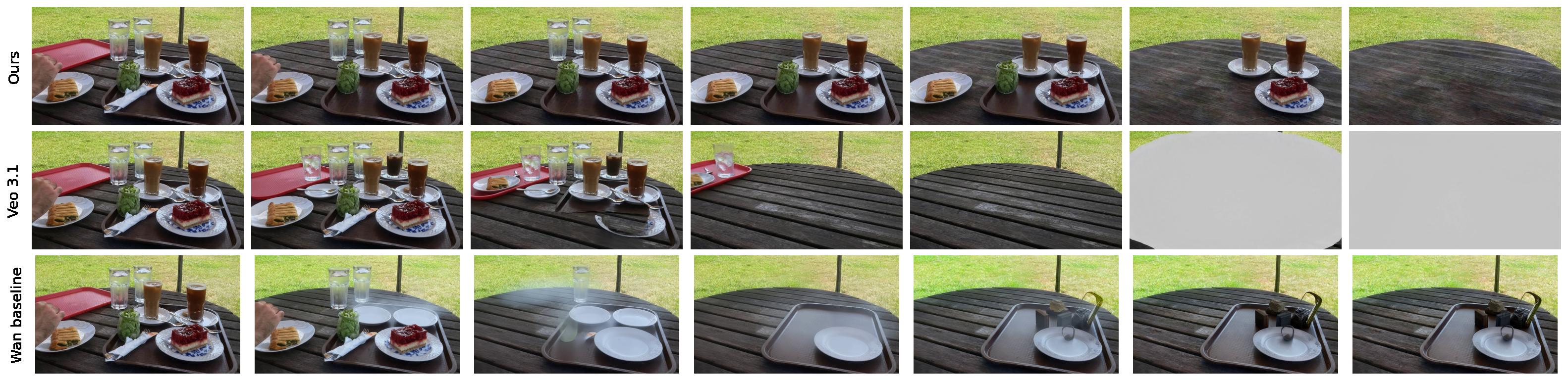}
  \vspace{0.4mm}

    \includegraphics[width=0.95\linewidth]{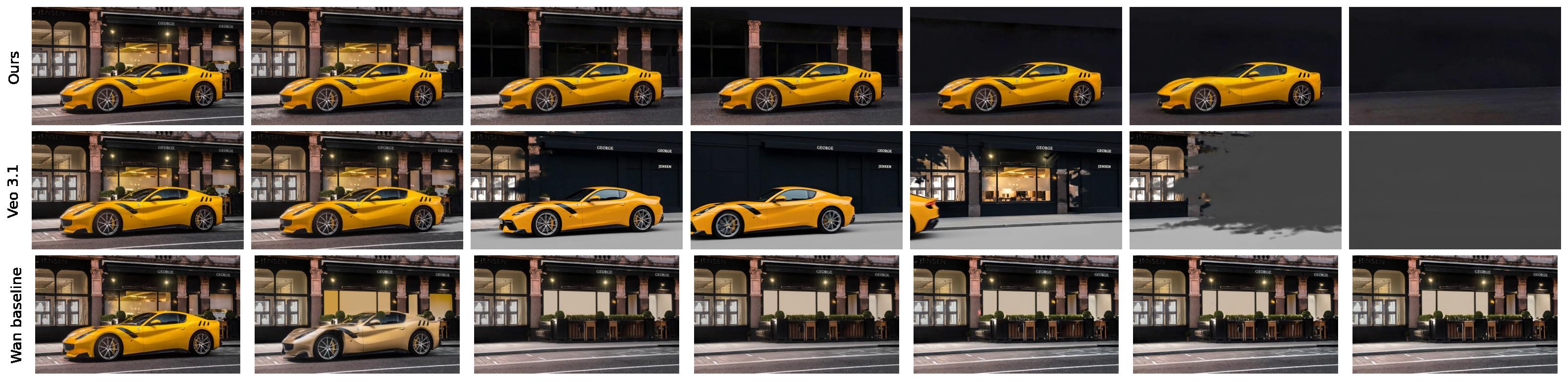}
  \vspace{0.4mm}

    \includegraphics[width=0.95\linewidth]{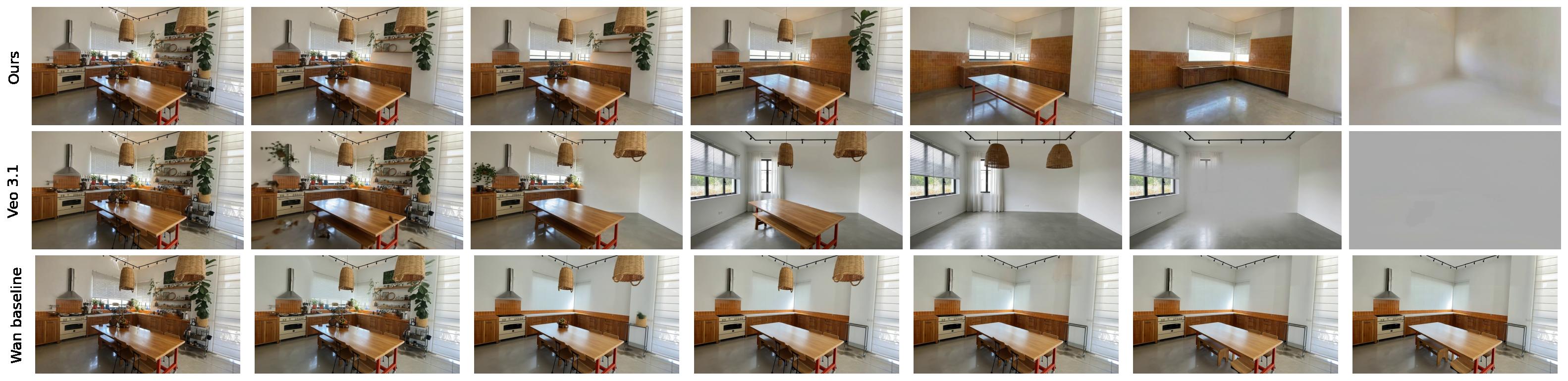}

  \vspace{0.4mm}
  
    \includegraphics[width=0.95\linewidth]{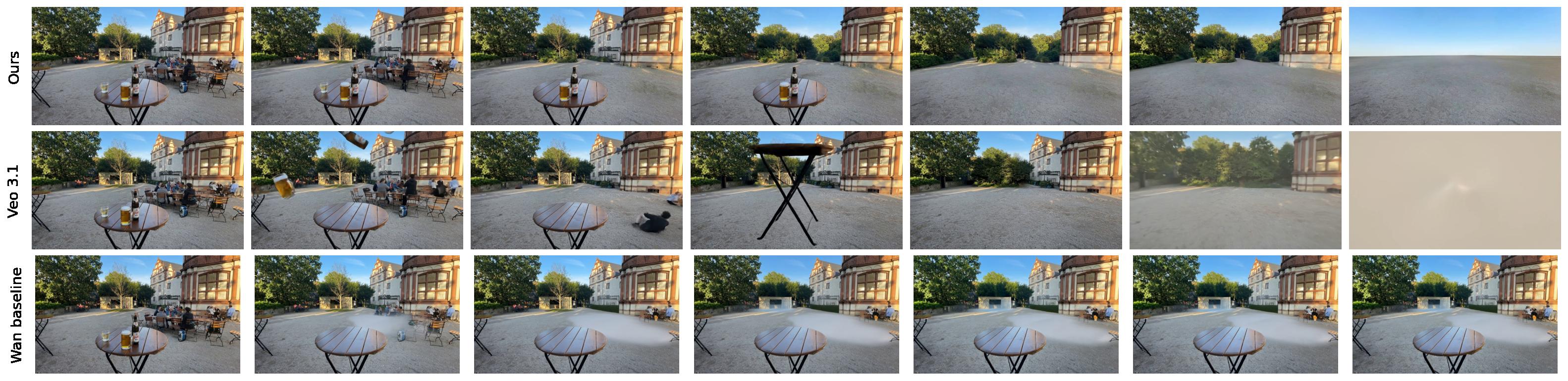}
  \vspace{0.4mm}
  
    \includegraphics[width=0.95\linewidth]{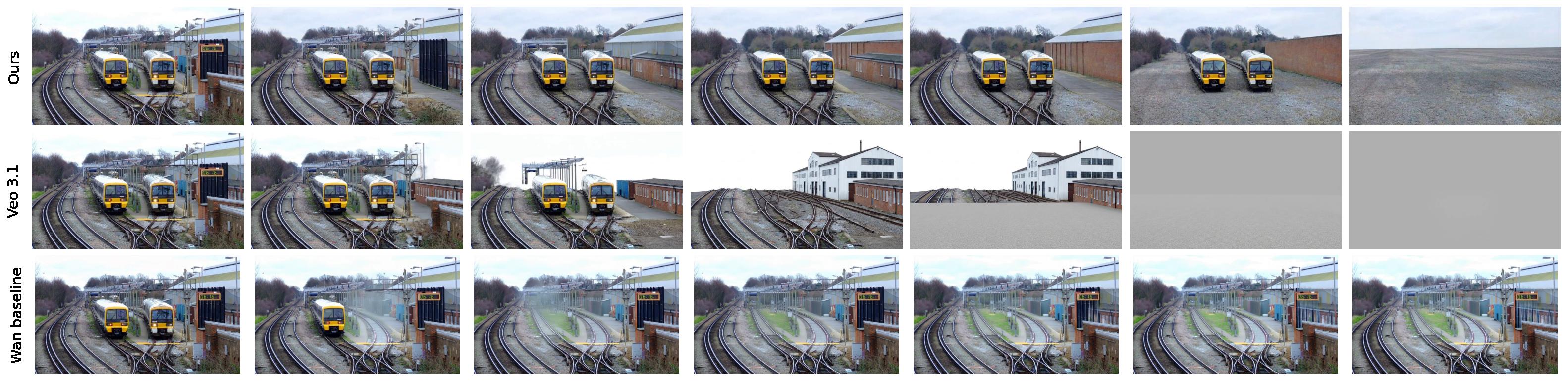}
  \caption{\textbf{Qualitative Comparison.} Each block shows (Top) \textbf{Ours}, (Middle) \textbf{Veo 3.1}, and (Bottom) \textbf{Wan 2.2}.}
  \label{fig:qualitative_part2}
\end{figure*}

\clearpage
\bibliographystyle{ACM-Reference-Format}
\bibliography{reference}

@String{Computing = "Computing" }

@String{Computer = "{IEEE} Computer" }

@String{Springer = "Springer-Verlag" }

@article{chen2007real,
  title={Real-time edge-aware image processing with the bilateral grid},
  author={Chen, Jiawen and Paris, Sylvain and Durand, Fr{\'e}do},
  journal={ACM Transactions on Graphics (TOG)},
  volume={26},
  number={3},
  pages={103--es},
  year={2007},
  publisher={Acm New York, NY, USA}
}

@article{farbman2008edge,
  title={Edge-preserving decompositions for multi-scale tone and detail manipulation},
  author={Farbman, Zeev and Fattal, Raanan and Lischinski, Dani and Szeliski, Richard},
  journal={ACM transactions on graphics (TOG)},
  volume={27},
  number={3},
  pages={1--10},
  year={2008},
  publisher={ACM New York, NY, USA}
}

@article{decarlo2002stylization,
  title={Stylization and abstraction of photographs},
  author={DeCarlo, Doug and Santella, Anthony},
  journal={ACM transactions on graphics (TOG)},
  volume={21},
  number={3},
  pages={769--776},
  year={2002},
  publisher={ACM New York, NY, USA}
}

@article{winnemoller2006real,
  title={Real-time video abstraction},
  author={Holger Winnem{\"o}ller and Sven C. Olsen and Bruce Gooch},
  journal={ACM Transactions On Graphics (TOG)},
  volume={25},
  number={3},
  pages={1221--1226},
  year={2006},
  publisher={ACM New York, NY, USA}
}

@inproceedings{hertzmann1998painterly,
  title={Painterly rendering with curved brush strokes of multiple sizes},
  author={Hertzmann, Aaron},
  booktitle={Proceedings of the 25th annual conference on Computer graphics and interactive techniques},
  pages={453--460},
  year={1998}
}

@article{Kang2009FlowBased,
  author={Henry Kang and Seungyong Lee and Charles K. Chui},
  journal={IEEE Transactions on Visualization and Computer Graphics},
  title={Flow-Based Image Abstraction},
  year={2009},
  volume={15},
  number={1},
  pages={62--76},
  doi={10.1109/TVCG.2008.81}
}

@inproceedings{Lu:2010:IPS,
 author = {Jingwan Lu and Pedro V. Sander and Adam Finkelstein},
 title = {Interactive Painterly Stylization of Images, Videos and {3D} Animations},
 booktitle = {Proceedings of ACM SIGGRAPH symposium on Interactive 3D Graphics and Games},
 year = {2010},
 month = feb,
}

@inproceedings{gatys2016image,
  title={Image style transfer using convolutional neural networks},
  author={Gatys, Leon A and Ecker, Alexander S and Bethge, Matthias},
  booktitle={Proceedings of the IEEE conference on computer vision and pattern recognition},
  pages={2414--2423},
  year={2016}
}

@article{vinker2022clipasso,
  title={Clipasso: Semantically-aware object sketching},
  author={Vinker, Yael and Pajouheshgar, Ehsan and Bo, Jessica Y and Bachmann, Roman Christian and Bermano, Amit Haim and Cohen-Or, Daniel and Zamir, Amir and Shamir, Ariel},
  journal={ACM Transactions on Graphics (TOG)},
  volume={41},
  number={4},
  pages={1--11},
  year={2022},
  publisher={ACM New York, NY, USA}
}

@inproceedings{vinker2023clipascene,
  title={Clipascene: Scene sketching with different types and levels of abstraction},
  author={Vinker, Yael and Alaluf, Yuval and Cohen-Or, Daniel and Shamir, Ariel},
  booktitle={Proceedings of the IEEE/CVF International Conference on Computer Vision},
  pages={4146--4156},
  year={2023}
}

@article{criminisi2004region,
  title={Region filling and object removal by exemplar-based image inpainting},
  author={Criminisi, Antonio and P{\'e}rez, Patrick and Toyama, Kentaro},
  journal={IEEE Transactions on image processing},
  volume={13},
  number={9},
  pages={1200--1212},
  year={2004},
  publisher={IEEE}
}

@inproceedings{kirillov2023segment,
  title={Segment anything},
  author={Kirillov, Alexander and Mintun, Eric and Ravi, Nikhila and Mao, Hanzi and Rolland, Chloe and Gustafson, Laura and Xiao, Tete and Whitehead, Spencer and Berg, Alexander C and Lo, Wan-Yen and others},
  booktitle={Proceedings of the IEEE/CVF international conference on computer vision},
  pages={4015--4026},
  year={2023}
}

@article{hu2022lora,
  title={Lora: Low-rank adaptation of large language models.},
  author={Hu, Edward J and Shen, Yelong and Wallis, Phillip and Allen-Zhu, Zeyuan and Li, Yuanzhi and Wang, Shean and Wang, Lu and Chen, Weizhu and others},
  journal={ICLR},
  volume={1},
  number={2},
  pages={3},
  year={2022}
}

@article{Winnemoeller:2006:RTV,
  author = {Winnem{\"o}ller, Holger and Olsen, Sven C. and Gooch, Bruce},
  title = {Real-Time Video Abstraction},
  journal = {ACM Trans. Graph.},
  issue_date = {July 2006},
  volume = {25},
  number = {3},
  pages = {1221--1226},
  articleno = {101},
  year = {2006},
}

@article{XDOG-2012,
author = {Winnem\"{o}ller, Holger and Kyprianidis, Jan Eric and Olsen, Sven C.},
title = {XDoG: An eXtended difference-of-Gaussians compendium including advanced image stylization},
year = {2012},
volume = {36},
number = {6},
journal = {Comput. Graph.},
pages = {740–753},
}

@inproceedings{suvorov2022resolution,
  title={Resolution-robust large mask inpainting with fourier convolutions},
  author={Suvorov, Roman and Logacheva, Elizaveta and Mashikhin, Anton and Remizova, Anastasia and Ashukha, Arsenii and Silvestrov, Aleksei and Kong, Naejin and Goka, Harshith and Park, Kiwoong and Lempitsky, Victor},
  booktitle={Proceedings of the IEEE/CVF winter conference on applications of computer vision},
  pages={2149--2159},
  year={2022}
}

@inproceedings{aberman2022deep,
  title={Deep saliency prior for reducing visual distraction},
  author={Aberman, Kfir and He, Junfeng and Gandelsman, Yossi and Mosseri, Inbar and Jacobs, David E and Kohlhoff, Kai and Pritch, Yael and Rubinstein, Michael},
  booktitle={Proceedings of the IEEE/CVF Conference on Computer Vision and Pattern Recognition},
  pages={19851--19860},
  year={2022}
}

@inproceedings{huynh2023simpson,
  title={SimpSON: Simplifying scenes by object neutralization},
  author={Huynh, Cong Duy and others},
  booktitle={Proceedings of the IEEE/CVF Winter Conference on Applications of Computer Vision},
  year={2023}
}

@article{bhattad2024visual,
  title={Visual Jenga: Object Removal via Counterfactual Inpainting},
  author={Bhattad, Anand and others},
  journal={arXiv preprint arXiv:2401.XXXX},
  year={2024}
}

@article{xing2023diffsketcher,
  title={Diffsketcher: Text guided vector sketch synthesis through latent diffusion models},
  author={Xing, Ximing and Wang, Chuang and Zhou, Haitao and Zhang, Jing and Yu, Qian and Xu, Dong},
  journal={Advances in Neural Information Processing Systems},
  volume={36},
  pages={15869--15889},
  year={2023}
}

@inproceedings{winter2024objectdrop,
  title={Objectdrop: Bootstrapping counterfactuals for photorealistic object removal and insertion},
  author={Winter, Daniel and Cohen, Matan and Fruchter, Shlomi and Pritch, Yael and Rav-Acha, Alex and Hoshen, Yedid},
  booktitle={European Conference on Computer Vision},
  pages={112--129},
  year={2024},
  organization={Springer}
}

@online{nano-banana,
    author = {Sharon, David and Brichtova, Nicole},
    title = {Image editing in Gemini just got a major upgrade},
    year = {2025},
    url = {https://blog.google/products/gemini/updated-image-editing-model} 
}

@inproceedings{fried2015finding,
  title={Finding distractors in images},
  author={Fried, Ohad and Shechtman, Eli and Goldman, Dan B and Finkelstein, Adam},
  booktitle={Proceedings of the IEEE Conference on Computer Vision and pattern Recognition},
  pages={1703--1712},
  year={2015}
}

@article{OpenImages,
  author = {Alina Kuznetsova and Hassan Rom and Neil Alldrin and Jasper Uijlings and Ivan Krasin and Jordi Pont-Tuset and Shahab Kamali and Stefan Popov and Matteo Malloci and Alexander Kolesnikov and Tom Duerig and Vittorio Ferrari},
  title = {The Open Images Dataset V4: Unified image classification, object detection, and visual relationship detection at scale},
  year = {2020},
  journal = {IJCV}
}

@article{wan2025,
      title={Wan: Open and Advanced Large-Scale Video Generative Models}, 
      author={Team Wan and Ang Wang and Baole Ai and Bin Wen and Chaojie Mao and Chen-Wei Xie and Di Chen and Feiwu Yu and Haiming Zhao and Jianxiao Yang and Jianyuan Zeng and Jiayu Wang and Jingfeng Zhang and Jingren Zhou and Jinkai Wang and Jixuan Chen and Kai Zhu and Kang Zhao and Keyu Yan and Lianghua Huang and Mengyang Feng and Ningyi Zhang and Pandeng Li and Pingyu Wu and Ruihang Chu and Ruili Feng and Shiwei Zhang and Siyang Sun and Tao Fang and Tianxing Wang and Tianyi Gui and Tingyu Weng and Tong Shen and Wei Lin and Wei Wang and Wei Wang and Wenmeng Zhou and Wente Wang and Wenting Shen and Wenyuan Yu and Xianzhong Shi and Xiaoming Huang and Xin Xu and Yan Kou and Yangyu Lv and Yifei Li and Yijing Liu and Yiming Wang and Yingya Zhang and Yitong Huang and Yong Li and You Wu and Yu Liu and Yulin Pan and Yun Zheng and Yuntao Hong and Yupeng Shi and Yutong Feng and Zeyinzi Jiang and Zhen Han and Zhi-Fan Wu and Ziyu Liu},
      journal = {arXiv preprint arXiv:2503.20314},
      year={2025}
}

@article{wiedemer2025video,
  title={Video models are zero-shot learners and reasoners},
  author={Wiedemer, Thadd{\"a}us and Li, Yuxuan and Vicol, Paul and Gu, Shixiang Shane and Matarese, Nick and Swersky, Kevin and Kim, Been and Jaini, Priyank and Geirhos, Robert},
  journal={arXiv preprint arXiv:2509.20328},
  year={2025}
}

@article{zhang2025image,
  title={Are Image-to-Video Models Good Zero-Shot Image Editors?},
  author={Zhang, Zechuan and Chen, Zhenyuan and Yang, Zongxin and Yang, Yi},
  journal={arXiv preprint arXiv:2511.19435},
  year={2025}
}

@article{wu2025chronoedit,
    title={ChronoEdit: Towards Temporal Reasoning for Image Editing and World Simulation},
    author={Wu, Jay Zhangjie and Ren, Xuanchi and Shen, Tianchang and Cao, Tianshi and He, Kai and Lu, Yifan and Gao, Ruiyuan and Xie, Enze and Lan, Shiyi and Alvarez, Jose M. and Gao, Jun and Fidler, Sanja and Wang, Zian and Ling, Huan},
    journal={arXiv preprint arXiv:2510.04290},
    year={2025}
}

@article{kendall1938new,
  title={A new measure of rank correlation},
  author={Kendall, Maurice G},
  journal={Biometrika},
  volume={30},
  number={1-2},
  pages={81--93},
  year={1938},
  publisher={Oxford University Press}
}

@book{kendall1948rank,
  title={Rank Correlation Methods},
  author={Kendall, Maurice G},
  year={1948},
  publisher={Charles Griffin \& Co.},
  address={London}
}

@article{zhang2021image,
  title={Image Composition Assessment with Saliency-augmented Multi-pattern Pooling},
  author={Zhang, Bo and Niu, Li and Zhang, Liqing},
  journal={arXiv preprint arXiv:2104.03133},
  year={2021}
}

@book{gombrich1960art,
  title     = {Art and Illusion: A Study in the Psychology of Pictorial Representation},
  author    = {Gombrich, Ernst H.},
  year      = {1960},
  publisher = {Princeton University Press}
}

@book{marr2010vision,
  title     = {Vision: A Computational Investigation into the Human Representation and Processing of Visual Information},
  author    = {Marr, David},
  year      = {2010},
  publisher = {The MIT Press}
}

@book{livingstone2002vision,
  title     = {Vision and Art: The Biology of Seeing},
  author    = {Livingstone, Margaret},
  year      = {2008},
  publisher = {Harry N. Abrams}
}

@book{palmer1999vision,
  title     = {Vision Science: Photons to Phenomenology},
  author    = {Palmer, Stephen E.},
  year      = {1999},
  publisher = {MIT Press}
}

@book{kemp1990science,
  title     = {The Science of Art: Optical Themes in Western Art from Brunelleschi to Seurat},
  author    = {Kemp, Martin},
  year      = {1992},
  publisher = {Yale University Press}
}

@inproceedings{wertheimer1938laws,
    author = {Wertheimer, Max},
    title = {Laws of Organization in Perceptual Forms},
    booktitle = {A source book of Gestalt psychology},
    pages   = {71--88},
    year = {1938}
}

@inproceedings{winkenbach1994computer,
  title={Computer-generated pen-and-ink illustration},
  author={Winkenbach, Georges and Salesin, David H},
  booktitle={Proceedings of the 21st annual conference on Computer graphics and interactive techniques},
  pages={91--100},
  year={1994}
}

@inproceedings{curtis1997computer,
  title={Computer-generated watercolor},
  author={Curtis, Cassidy J and Anderson, Sean E and Seims, Joshua E and Fleischer, Kurt W and Salesin, David H},
  booktitle={Proceedings of the 24th annual conference on Computer graphics and interactive techniques},
  pages={421--430},
  year={1997}
}

@inproceedings{10.1145/383259.383295,
author = {Hertzmann, Aaron and Jacobs, Charles E. and Oliver, Nuria and Curless, Brian and Salesin, David H.},
title = {Image analogies},
year = {2001},
isbn = {158113374X},
publisher = {Association for Computing Machinery},
address = {New York, NY, USA},
url = {https://doi.org/10.1145/383259.383295},
doi = {10.1145/383259.383295},
booktitle = {Proceedings of the 28th Annual Conference on Computer Graphics and Interactive Techniques},
pages = {327–340},
numpages = {14},
series = {SIGGRAPH '01}
}

@article{willats2005defining,
  title={Defining pictorial style: Lessons from linguistics and computer graphics},
  author={Willats, John and Durand, Fr{\'e}do},
  journal={Axiomathes},
  volume={15},
  number={3},
  pages={319--351},
  year={2005},
  publisher={Springer}
}

@inproceedings{10.1145/508530.508550,
author = {Durand, Fr\'{e}do},
title = {An invitation to discuss computer depiction},
year = {2002},
isbn = {1581134940},
publisher = {Association for Computing Machinery},
address = {New York, NY, USA},
url = {https://doi.org/10.1145/508530.508550},
doi = {10.1145/508530.508550},
booktitle = {Proceedings of the 2nd International Symposium on Non-Photorealistic Animation and Rendering},
pages = {111–124},
numpages = {14},
location = {Annecy, France},
series = {NPAR '02}
}

\appendix

\section{Classifier Training}
\label{sec:classifier}
To train the verification classifier used in Stage I, we constructed a balanced dataset of approximately 8,000 image pairs (original and edited crops). The primary objective of this classifier is to evaluate the removal quality.

\paragraph{Data Construction and Labeling Criteria.}
The dataset consists of positive and negative samples, labeled based on strict visual fidelity and accurate element removal:
\begin{itemize}
    \item \textbf{Positive Samples ($\sim$4,000 pairs):} These were manually curated from high-quality removal trajectories. A pair was labeled as positive only if the target object was entirely removed without leaving residual artifacts (e.g., floating shadows) and the infilled background maintained strict photorealism consistent with the surrounding scene.
    \item \textbf{Negative Samples ($\sim$4,000 pairs):} To ensure the classifier is robust against various failure modes, negative samples were gathered using three distinct strategies:
    \begin{enumerate}
        \item \textit{Manual Pipeline Failures:} Manually selected examples of failed automated edits, such as incomplete removals, hallucinated objects, or noticeable boundary/blending artifacts.
        \item \textit{Reversed Trajectories:} We reversed the chronological order of positive samples, presenting an object \textit{insertion} to the network. This teaches the model to penalize the appearance of new, unprompted elements when a removal was expected.
        \item \textit{Synthetic Adversarial Edits:} We explicitly prompted the generative model to produce unwanted modifications, such as altering the background style, replacing the target element instead of removing it, or severely distorting the surrounding context.
    \end{enumerate}
\end{itemize}

\begin{figure}[htbp]
  \centering
  
  \subfloat[Normalized Confusion Matrix]{
    \includegraphics[width=\linewidth]{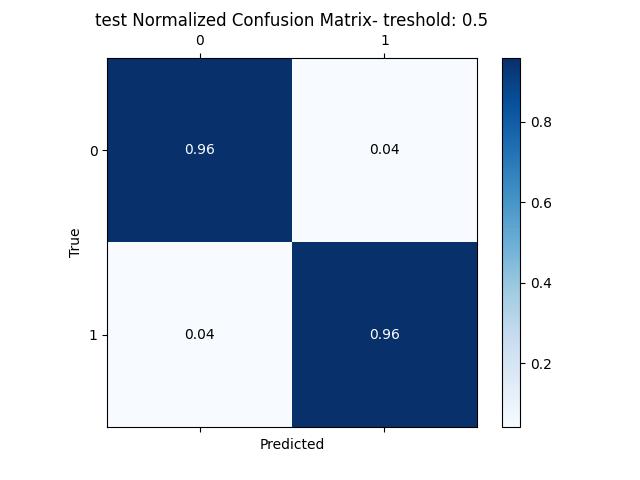}
  }
  
  \vspace{1em} 
  
  \subfloat[Train and Validation Loss and Validation Accuracy over Epochs]{
    \includegraphics[width=\linewidth]{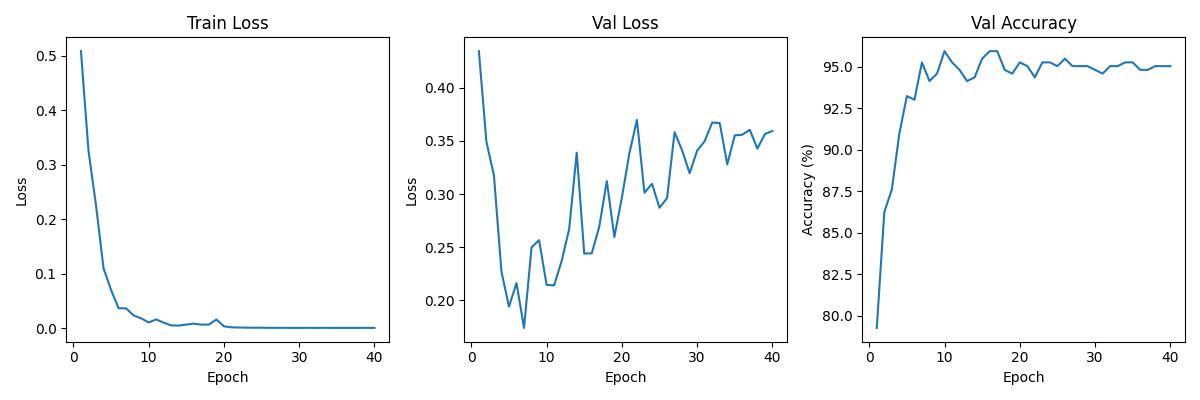}
  }
  
  \caption{\textbf{Classifier Performance.} (a) The model achieves 96\% True Positive and True Negative rates at the decision threshold of 0.5. (b) Validation accuracy stabilizes after approximately 10 epochs.}
  \label{fig:classifier_metrics}
\end{figure}

\paragraph{Architecture and Training Details.}
As noted in the main text, we employ a Siamese network architecture based on a pre-trained DINOv2-Large backbone. To maintain training efficiency, we inject Low-Rank Adaptation (LoRA) modules ($r=16$) into the attention and feed-forward layers (Sec. 4.1 on the main paper).
In Fig.~\ref{fig:classifier_metrics}, we present the training progression and final confusion matrix. The final model was fine-tuned for 16 epochs. The validation accuracy plateaus at approximately 96\%, with the confusion matrix confirming a balanced performance between accepting valid edits and rejecting artifacts.

\section{Taxonomy Human Studies}

To verify that the taxonomy defined in Section~3.1 aligns with human perception, we conducted a study with 93 images annotated by three independent raters. We selected elements for removal, and asked the raters, which are image quality experts, to classify the object based on our taxonomy.
We calculated Kendall's Rank Correlation Coefficient ($\tau_b$), utilizing it instead of standard metrics because our data is \textit{ordinal} and contains frequent \textit{ties}. The pairwise agreement scores between $3$ pairs of raters were: $0.75, 0.63, 0.65$. To analyze disagreements, we computed a normalized confusion matrix (Fig.~\ref{fig:combined_analysis}). Raters showed high consistency on the most frequent category (``Secondary/Contextual'': 83.2\%). Disagreements were largely confined to boundary layers, most notably between ``Secondary'' and ``Background'' (32.1\%), reflecting the inherent ambiguity of where specific context ends and the general environment begins. Crucially, confusion between the hierarchy's extremes (``Primary'' vs. ``Background'') was negligible ($<2.5\%$), confirming that the core layers of our simplification model are semantically distinct.

Beyond static classification, we evaluated the robustness of the implied removal order by asking raters to choose which of two elements should be removed first, providing a ``Doesn't matter / Equal priority'' option to capture inherent subjectivity. The results illustrate why a coarse-grained taxonomy is a necessary scaffold to simplify the abstraction problem, serving as a robust approximation where granular human consensus fails. When comparing elements belonging to the \textit{same} semantic category (e.g., two secondary objects; $N \approx 1600$ pairs), raters indicated the order did not matter in 27.5\% of cases, confirming high ambiguity at this granular level. 

Conversely, when comparing elements belonging to \textit{different} semantic categories ($N \approx 500$ pairs), the ``doesn't matter'' rate dropped significantly to only 3.7\%. When a preference was expressed in these inter-category pairs, the direction was highly consistent with our hierarchy, as detailed in Table~1. The results show a remarkable consensus: ``Distractor'' elements were universally chosen for removal before any other category (100\%). Similarly, ``Secondary'' elements were prioritized over ``Primary'' subjects in 96.1\% of decisive comparisons. The dominant values in the upper triangle of the table indicate a strong, unidirectional ordering that strongly validates our taxonomy as a robust high-level guide for abstraction.

\section{Automated Object Removal Detection Pipeline}

We developed an automated pipeline to identify the specific frame $t^*$ where a ground-truth object $O$ is removed. Let $M \in \{0, 1\}^{H \times W}$ be the binary mask of the object.
We analyze the sequence of generated frames $\{F_1, \dots, F_N\}$ by first computing a binary difference mask $D_t$ for each time step, derived from optical flow and thresholding (as detailed in Sec. 3 in the paper). We define the \textbf{cumulative difference mask} $C_t$ as the logical disjunction of changes up to time $t$:
\begin{equation}
    C_t = \bigcup_{k=1}^t D_k.
\end{equation}

The removal frame $t^*$ is identified as the first time step satisfying two concurrent conditions:

\begin{enumerate}
    \item \textit{Cumulative Coverage:} We ensure the removal is complete by verifying that accumulated changes sufficiently cover the object area:
    \begin{equation}
        \frac{|C_t \cap M|}{|M|} \geq \tau_{\text{cov}}.
    \end{equation}
    We used $\tau_{\text{cov}} = 0.4$

    \item \textit{Spatiotemporal Locality:} To distinguish a specific removal event from accumulated noise, we analyze a local aggregate mask $L_t$ within a temporal window $[t-w, t+w]$. We enforce that significant changes must occur \textit{inside} the object region while the background remains stable:
    \begin{equation}
        \underbrace{ \frac{| L_t \cap M |}{| M |} \geq \tau_{\text{act}} }_{\text{Active Removal}}
        \qquad \text{and} \qquad
        \underbrace{ \frac{| L_t \setminus M |}{| M^c |} \leq \tau_{\text{stab}} }_{\text{Background Stability}},
    \end{equation}
    where $M^c$ denotes the background region. We used $\tau_{\text{act}} = 0.4$, and $\tau_{\text{stab}} = 0.1$.
\end{enumerate}

\section{Manual Data Filtering.}
For our distilled image-to-video model, we found that it is critical to use high-quality training samples. Because the automated validation occasionally yielded false positives, introducing noise into the training signal (Sec. 4 of the main paper and Sec.~\ref{sec:classifier}), we performed a manual review to retain only the highest quality samples. Viewing the generation as a Markovian process where each step relies strictly on its immediate predecessor, we intervened to preserve visual continuity. Specifically, we split sequences that violated this continuity via abrupt or excessively large changes, removed specific noisy frames, and entirely discarded trajectories that suffered from irrecoverable degradation.

\section{Prompt Details for Search-Based Trajectory}

In this section, we provide the exact prompts used to drive the Vision–Language Model (VLM) planner and the generative inpainting model during Stage I (Search-Based Trajectory Generation). The system relies on three distinct types of prompts: (1) \textbf{Object Enumeration} to populate the list of candidates based on our taxonomy; (2) \textbf{Object Selection} to choose the specific element to remove; and (3) \textbf{Inpainting Instructions} to guide the generative erasure.

Note: Text enclosed in braces (e.g., \texttt{\{list\_of\_objects\}}) represents placeholders populated dynamically at inference time.

\subsection{Object Enumeration Prompts}
The following prompts are used to identify candidate objects in the scene. The specific prompt selected depends on the current active category level in the taxonomy.

\paragraph{A. Taxonomy Level: Distracting / Unwanted Elements}
\begin{quote}
\ttfamily
\small
Act as an expert visual analyst for a high-end photo editing studio. Your mission is to autonomously identify a photograph's primary subject and then conduct a comprehensive audit to list every possible element that detracts from it.

\textbf{1. Core Mission: Subject-First Analysis}\\
Your entire analysis is driven by one question: What is the main subject, and what is distracting from it? You must first identify the most likely primary subject(s) based on composition, focus, and human interest. Then, you will identify all other elements that are non-essential, distracting, or diminish the visual impact of that subject.

\textbf{2. Core Instructions \& Mindset}\\
Step 1: Identify the Primary Subject. Analyze the image and determine its most logical subject(s). This could be a person, a group of people, a specific object, or an architectural feature.\\
Step 2: Perform an Exhaustive Scan for Distractions. Based on your identified subject, your goal now is to be meticulous and find a large number of items that pull focus away from it. Do not stop at the most obvious distractions; look for minor ones as well.\\
Step 3: List All Distracting Elements. For each element, provide a precise description and its location. Explain why it distracts from the primary subject you identified.

\textbf{3. Principles for Exhaustive Identification}\\
Your default is to be over-inclusive. When in doubt, list the object. The editor will make the final decision. Use this checklist to find elements that distract from the main subject:
\begin{itemize}
    \item Litter \& Debris: Systematically scan for any man-made trash.
    \item Unwanted Objects: Identify items that feel out of place in the context of the subject (e.g., modern signs near a historic monument, power lines cluttering a portrait background).
    \item Temporary or Accidental Items: Look for things that are not part of the intended scene (e.g., a stray backpack, a coffee cup, equipment).
    \item Background People \& Vehicles: Identify any person or vehicle that is clearly not the primary subject. List them individually as they compete for attention.
    \item Foreground Occlusions: Pay extremely close attention to the area in front of and around the main subject. List anything that partially blocks or intrudes upon its silhouette (e.g., a foreground tree branch, a lamppost).
\end{itemize}

\textbf{4. Important Specific Exclusions}\\
Do not list body parts (e.g., 'hand', 'face', 'leg'). Refer to people as a whole entity.\\
Do not list clothing items worn by any person (except for distracting or out-of-place accessories).\\
Do not list the primary subject(s) you identified in Step 1.\\
Do not list large, essential background elements (e.g., 'the sky', 'a wall', "grass", "vegetation"). Instead, list specific unwanted features on them (e.g., 'graffiti on the wall').

\textbf{5. Output Format}\\
Your response MUST be a single, machine-readable JSON object. Do not include any text or explanations outside of the JSON structure. The JSON object must contain two keys: "primary\_subject" (a string) and "list\_objects" (a list of lists, where each inner list is [object\_name, description]).
\end{quote}

\paragraph{B. Taxonomy Level: Structural / Amorphous Elements}
\begin{quote}
\ttfamily
\small
Act as an expert in visual scene analysis. Your task is to analyze the provided image and identify the main structural components and large surfaces that form the environment.

Focus on elements like walls, floors, ceilings, and baseboards. Do not list small, discrete objects like furniture or decorations. For each identified component, provide a brief description, including its color and material if discernible.

\textbf{Important Specific Exclusions:}\\
* Do not list body parts: Avoid itemizing individual body parts such as 'hand', 'leg', or 'face'. If a person is relevant to the scene, they can be described as a whole entity (e.g., 'a woman reading a book').\\
* Do not list clothing: Avoid itemizing individual articles of clothing like 'shirt', 'pants', or 'swimsuit'. These items are considered part of a person's appearance, not separate objects for this analysis.\\
* You \textbf{can} identify and list distinct accessories like hat, bag, sunglasses, handbags, watches, or jewelry.

\textbf{Output Format:}\\
Your response MUST be a single, machine-readable JSON object. Do not include any text or explanations outside of the JSON structure. The JSON object must contain two keys: "primary\_subject" (a string) and "list\_objects" (a list of lists, where each inner list is [object\_name, description]).
\end{quote}

\paragraph{C. Taxonomy Level: General / Other}
\begin{quote}
\ttfamily
\small
Act as a visual scene deconstruction engine. Analyze the provided image and return a comprehensive detailed list of all \{p\_level\} elements in the scene.

For each identified item, provide a description.

\textbf{Important Specific Exclusions:}\\
* Do not list body parts: Avoid itemizing individual body parts such as 'hand', 'leg', or 'face'. If a person is relevant to the scene, they can be described as a whole entity.\\
* Do not list clothing: Avoid itemizing individual articles of clothing like 'shirt', 'pants', or 'swimsuit'. These items are considered part of a person's appearance, not separate objects for this analysis.\\
* You \textbf{can} identify and list distinct accessories like hat, bag, sunglasses, handbags, watches, or jewelry.\\
* Do not list large, essential background elements (e.g., 'the sky', 'a wall', "grass", "vegetation").

\textbf{Output Format:}\\
Your response MUST be a single, machine-readable JSON object. Do not include any text or explanations outside of the JSON structure. The JSON object must contain two keys: "primary\_subject" (a string) and "list\_objects" (a list of lists, where each inner list is [object\_name, description]).
\end{quote}

\subsection{Element Selection Prompts}
Once a list of candidates is generated, the following prompts are used to select the specific object to remove at iteration $k$.

\paragraph{For Distracting Elements}
\begin{quote}
\ttfamily
\small
You are an intelligent image analysis agent. Your mission is to select the single most appropriate object to remove from an image, based on a provided list and a strict priority system.

You will be given an image and the following list of identified objects / elements, where each object is a tuple prefixed with its unique index - (ID, name\_object, description).

Identify one object from the list that has the least importance or relevance to the overall aesthetic and composition of the given image so I will remove it.

Follow these rules for your selection:\\
\textbf{Prioritize Clutter:} Start by looking for out-of-place, man-made items. These are the most common distractors.\\
\textbf{Excellent choices:} Stray cables, logo, power outlets, plugs, forgotten phone chargers, price tags, stray wrappers, loose papers, abstraction, a single misplaced item on an otherwise clean surface.

Based on these rules, identify one object. Your response MUST be the single tuple (ID, name\_object, description) that corresponding to the object you selected. Do not include any other text, explanations, or formatting.

The given list is: \{list\_of\_objects\}
\end{quote}

\paragraph{For General Elements}
\begin{quote}
\ttfamily
\small
You are an intelligent image analysis agent. Your mission is to select the single most appropriate object to remove from an image, based on a provided list and a strict priority system.

You will be given an image and the following list of identified objects / elements, where each object is a tuple prefixed with its unique index - (ID, name\_object, description).

Identify one object from the list that has the least importance or relevance to the overall aesthetic and composition of the given image so I will remove it.
Your response must be the single tuple (ID, name\_object, description) that corresponding to the object you selected from the provided list. Do not include any other text, explanations, code blocks, or markdown formatting.

The given list is: \{list\_of\_objects\}
\end{quote}

\subsection{Inpainting Prompts}
The selected object is removed using a generative model. We utilize two variations of prompts to guide the inpainting process.

\paragraph{Variation 1 (Direct Removal)}
\begin{quote}
\ttfamily
\small
Remove from the given image only the \{OBJECT\}.
Remove all associated shadows and reflections of the this Object.
The reconstruction must be seamless and realistic. Keep the rest of the image the same - do not add any new or different objects.
The primary goal it to make the image less cluttered.
\end{quote}

\paragraph{Variation 2 (Abstractive Removal)}
\begin{quote}
\ttfamily
\small
With the primary goal of making the image less cluttered, generate an alternate version of this image with removing the \{OBJECT\}. The space it occupied, along with its shadows and reflections, should be realistically filled by continuing the natural lines, textures, and lighting of the surrounding environment. The final image must look cleaner, more organized.
\end{quote}

\section{Prompt Details for Baseline Comparisons}
To ensure a fair and consistent quantitative evaluation (as detailed in the main text), we used the exact same text prompt for all off-the-shelf image-to-video (I2V) baseline models (e.g., Veo 3.1 and Wan 2.2).

\paragraph{Video Generation Prompt for Baselines}
\begin{quote}
\ttfamily
\small
A photorealistic, stop-motion video with a static, locked-off camera. The video will show a progressive, subtractive abstraction of the scene through a specific sequence of object removals: (1) First, small clutter, litter, and distracting details pop out of existence. (2) Next, secondary furniture and context objects disappear. (3) Then, the main subject elements vanish. (4) Finally, the background amorphous structures dissolve, leaving a neutral void. As each element vanishes, the surrounding background must be seamlessly inpainted. Maintain the original lighting and perspective throughout.
\end{quote}

\section{Direct Prompting Baselines}
\label{sec:direct_prompting}

To evaluate the necessity of our proposed \emph{Select-Remove-Verify} pipeline, we investigated whether contemporary state-of-the-art vision-language and generative models (e.g., NanoBanana Pro, Gemini 3 Pro) could achieve progressive photorealistic abstraction through direct prompting alone. We tested two primary naive approaches: end-to-end sequence generation and iterative step-by-step prompting. Both approaches exhibited critical limitations that justify our systematic framework.

\subsection{End-to-End Generation Failure}
Subtractive abstraction of natural scenes, as defined in our work (removing one element per step to maintain plausibility), typically requires extensive trajectories (often more then $6$ steps). Because current models are constrained to short sequences (4--6 frames) and cannot natively generate full, prolonged trajectories, we attempted to condense the task. We instructed the model to generate a 4-frame sequence corresponding directly to the categorical progression of our taxonomy. We utilized the following comprehensive system prompt:

\begin{quote}
\ttfamily
\small
You are an image deconstruction engine. Your task is to generate a 4 separate frame sequence based on the provided input image ($x0$).

Consistency is Key: Every image in the sequence must maintain the identical camera angle, perspective, lighting, and overall style of the original source image.
Flawless Removal: The removal of objects must be perfect. The space they once occupied should be filled in so seamlessly that it looks like they were never there.
Separate Files: The output must be 4 individual, separate image files, not a GIF or a single composite image.

In each consecutive frame, you must remove specific categories of elements to progressively abstract the scene. Follow this strict removal sequence:

Frame 1 ($x1$) = Clean the Scene\\
Input: Original Image ($x0$)\\
Action: Remove "Distracting / Unwanted Elements".\\
Definition: Remove visual noise that adds no value. Target trash, photobombers, stray branches, excessive text, date stamps, and bright hotspots.\\
Goal: A pristine, balanced version of the original photo.

Frame 2 ($x2$) = Isolate the Subject\\
Input: Result of Frame 1 ($x1$)\\
Action: Remove "Contextual / Secondary Elements".\\
Definition: Remove elements that provide atmosphere or narrative but are not the main focus. Target supporting characters, side objects, crowds, or environmental clues.\\
Goal: The main subject stands alone in a clear environment.

Frame 3 ($x3$) = Remove the Hero\\
Input: Result of Frame 2 ($x2$)\\
Action: Remove "Primary Subject(s)".\\
Definition: Remove the main focal point or "hero" of the image (the person, animal, or object the photo was built around).\\
Goal: An "empty scene" composition; a stage without actors.

Frame 4 ($x4$) = Erase the Canvas\\
Input: Result of Frame 3 ($x3$)\\
Action: Remove "Background / Amorphous Elements".\\
Definition: Remove distinct textures and boundaries. Smooth out the sky, water, ground, walls, and patterns.\\
Goal: Pure minimalism. A blend of colors or a solid void.

Output Format:\\
Present the sequence as:\\
$x1$: $x0$ / (Distracting Elements Removed)\\
$x2$: $x1$ / (Contextual Elements Removed)\\
$x3$: $x2$ / (Primary Subject Removed)\\
$x4$: $x3$ / (Background Removed)\\
Given Image ($x0$):
\end{quote}

As illustrated in Figure~\ref{fig:unstable_sequences}, attempts to force the model to complete to  map a complex sequence end-to-end yielded unstable and unrealistic results. The models struggled to perfectly isolate elements, frequently hallucinating new structures, severely warping the remaining background, and failed to maintain the strict photorealism required for photographic abstraction.

\subsection{Iterative Prompting Degradation}
We also evaluated a sequential approach, feeding the output of one edit back into the model for the next step (Figure~\ref{fig:iterative_degradation}). We observed that step-by-step editing on current image editors rapidly degrades global image quality. Generative models tend to introduce subtle global shifts in lighting, alter unprompted background details, or leave visible artifacts at the removal sites. Without the essential guardrails provided by our method, specifically explicit localized masking, alignment, image blending, and an independent verification classifier to reject bad edits, these small errors accumulate exponentially. After just a few iterations, the model routinely loses strict photorealism. Our structured \emph{Select-Remove-Verify} pipeline overcomes these issues to extract high-quality, stable trajectories.

\section{Detailed Construction of Confusion Matrices}
\label{sec:confusion_matrix_details}

In Fig. 4 of the main text, we present two confusion matrices: an Inter-Rater disagreement matrix and a Gemini vs. Raters disagreement matrix. We detail their exact mathematical construction below.

\paragraph{Dataset.}
To ensure a rigorous and unbiased evaluation, both matrices were generated using the same dataset of elements extracted from the 93 test images.

\paragraph{Rater Expertise.}
The three independent raters who participated in the study are professional photographers. Their extensive expertise in visual arts, composition, subject isolation, and aesthetic decluttering makes them uniquely qualified to evaluate the semantic importance of scene elements. 

\paragraph{Inter-Rater Matrix Construction (Fig. 4, Left).}
The purpose of this matrix is to visualize the baseline ambiguity of human perception. Because all three human raters are considered equal, we cannot arbitrarily assign one rater to the rows (as "predictions") and another to the columns (as "ground truth"). Therefore, the matrix is constructed to be undirected.
\begin{enumerate}
\item \textbf{Pairwise Aggregation:} We first compute standard directional cross-tabulations for all three rater combinations (Rater 1 vs. Rater 2, Rater 2 vs. Rater 3, and Rater 1 vs. Rater 3).
\item \textbf{Symmetrization:} To remove any directional bias, we sum each pairwise matrix with its transpose ($M + M^T$). Consequently, the underlying aggregated matrix of \textit{raw counts} is perfectly symmetric.
\item \textbf{Row Normalization:} To meaningfully interpret these raw counts as disagreement rates, the final visualization is row-normalized. 
\end{enumerate}

\paragraph{Gemini vs. Raters Matrix Construction (Fig. 4, Right).}
Unlike the inter-rater matrix, this comparison is inherently directional, as we are evaluating a single model prediction against human baselines.
\begin{enumerate}
\item \textbf{Aggregation:} For every element in the dataset, Gemini's single predicted category is compared against the individual annotations of all three human raters. These comparisons are aggregated into a raw count matrix where the rows represent Gemini's prediction and the columns represent the human raters' assignments.
\item \textbf{Row Normalization:} This matrix is then row-normalized. The resulting percentages represent the conditional probability of human categorization given a specific Gemini prediction.
\end{enumerate}

\begin{figure*}[h] 
  \centering
    \begin{tabular}{cc}
    \includegraphics[width=0.48\linewidth, height=0.4\linewidth]{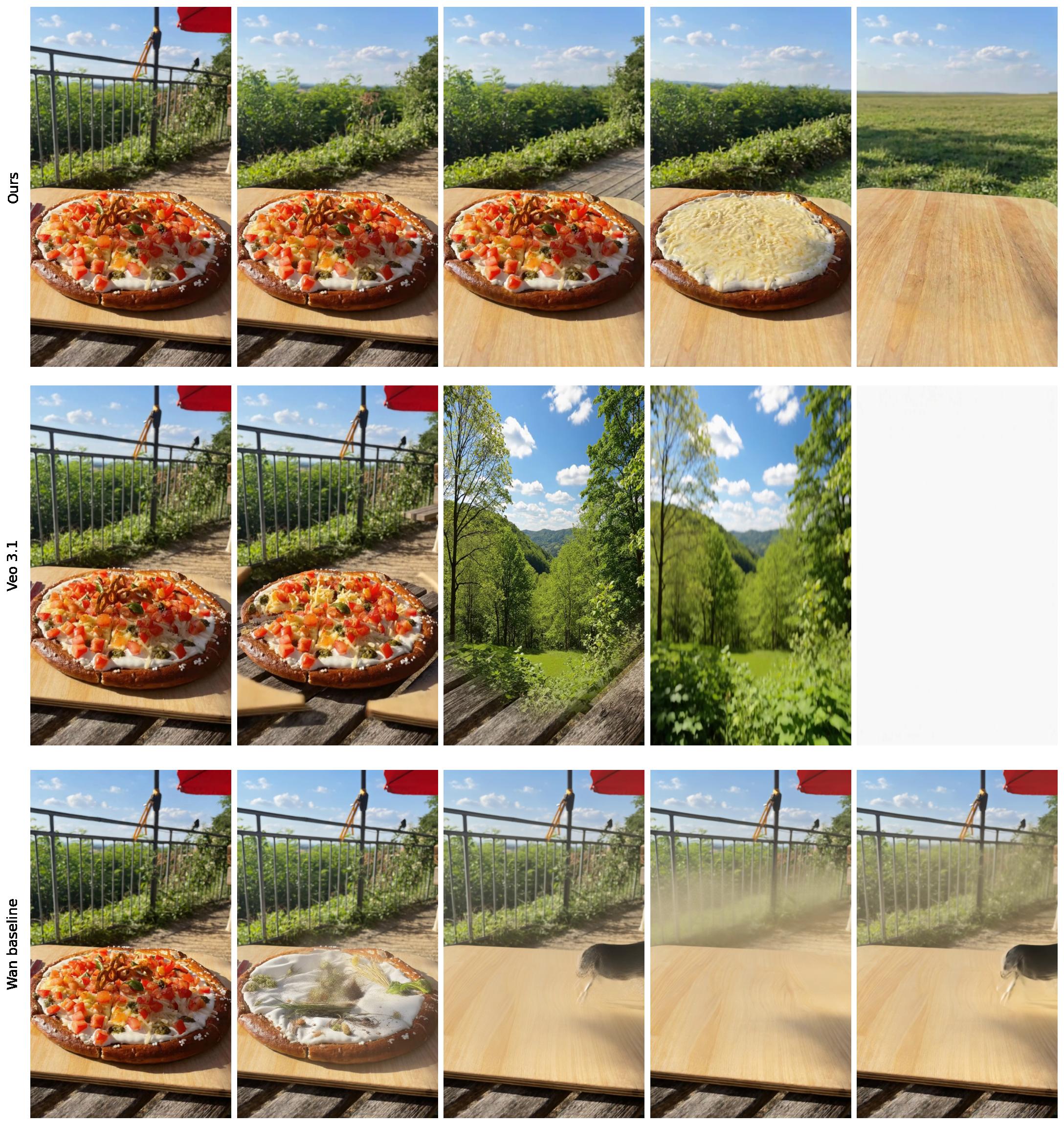} &
    \includegraphics[width=0.48\linewidth, height=0.4\linewidth]{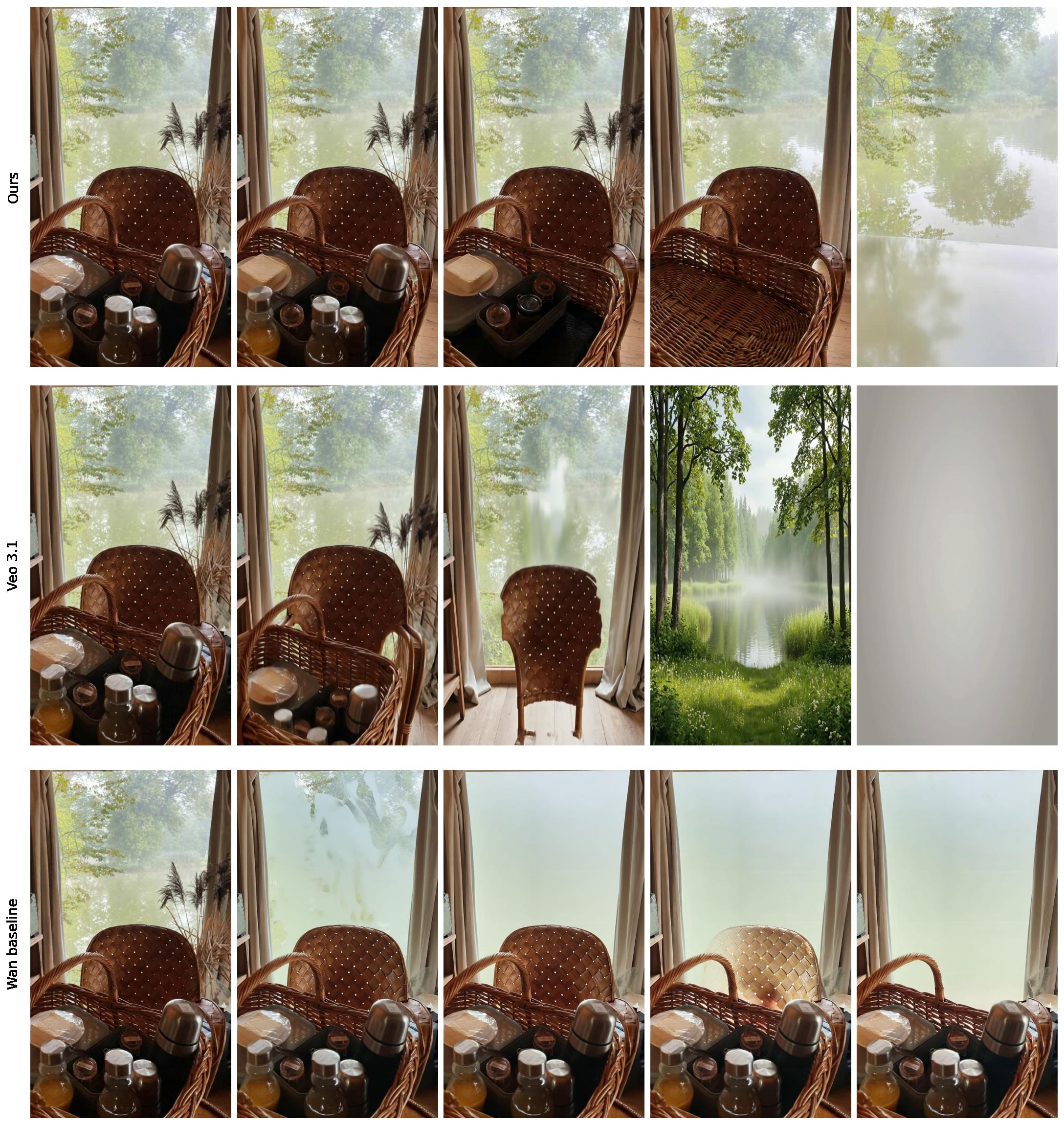} \\
    \end{tabular}
     
  \caption{\textbf{Qualitative Comparison.} Each block shows (Top) \textbf{Ours}, (Middle) \textbf{Veo 3.1}, and (Bottom) \textbf{Wan 2.2}.}
 \label{fig:qualitative_part1}
\end{figure*}

\begin{figure*}[t]
  \centering
  
  \includegraphics[width=\linewidth,height=0.13\linewidth]{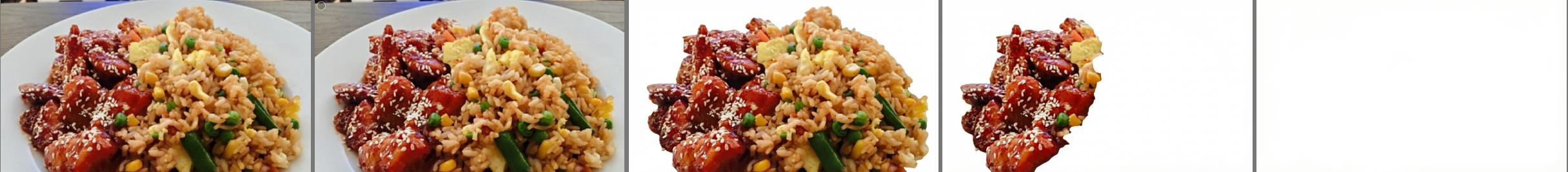} \\
  \vspace{0.1cm} 
  
  \includegraphics[width=\linewidth,height=0.13\linewidth]{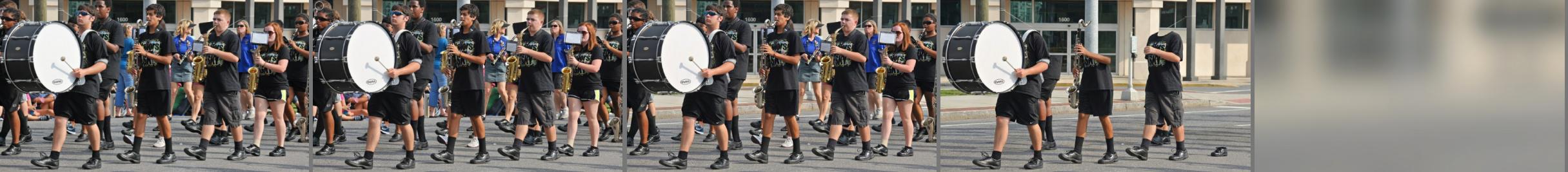} \\
  \vspace{0.1cm} 
  
  
  \includegraphics[width=\linewidth,height=0.13\linewidth]{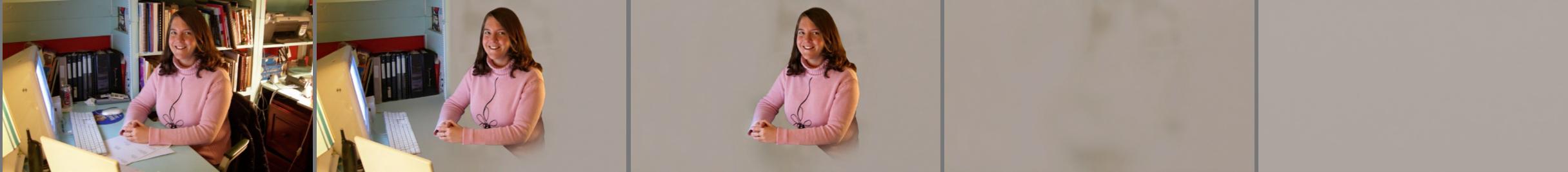} \\
  \vspace{0.1cm} 
  
  
  \includegraphics[width=\linewidth,height=0.13\linewidth]{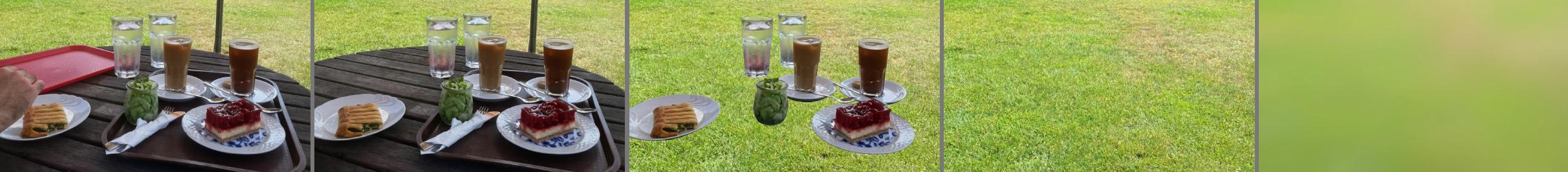} 

  \caption{\textbf{End-to-End Direct Prompting Failure.} When state-of-the-art generative models are prompted to generate a progressive abstraction sequence directly, the results are unstable. Each row illustrates a different example where the models fail to perfectly isolate the removals, hallucinate new structures, and progressively lose photorealism across the generated sequence.}
  \label{fig:unstable_sequences}
\end{figure*}

\begin{figure*}[t] 
  \centering 
\makebox[0.16\linewidth][c]{Before}%
 \makebox[0.16\linewidth][c]{After}\hfill
 \makebox[0.16\linewidth][c]{Before}%
 \makebox[0.16\linewidth][c]{After} \hfill
 \makebox[0.16\linewidth][c]{Before}%
 \makebox[0.16\linewidth][c]{After} \\
 \vspace{0.02cm}
  \includegraphics[width=0.31\linewidth, height=0.17\linewidth]{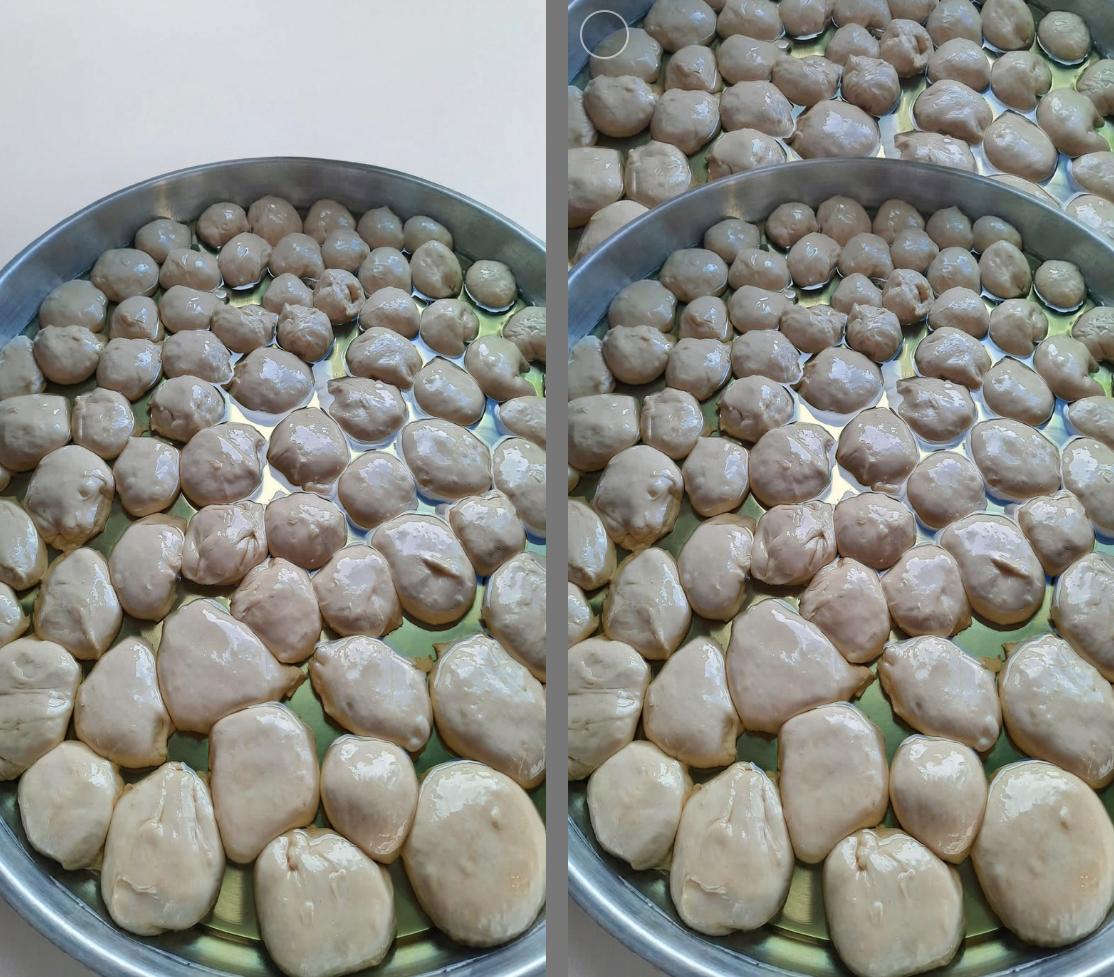} \hfill
  \includegraphics[width=0.31\linewidth, height=0.17\linewidth]{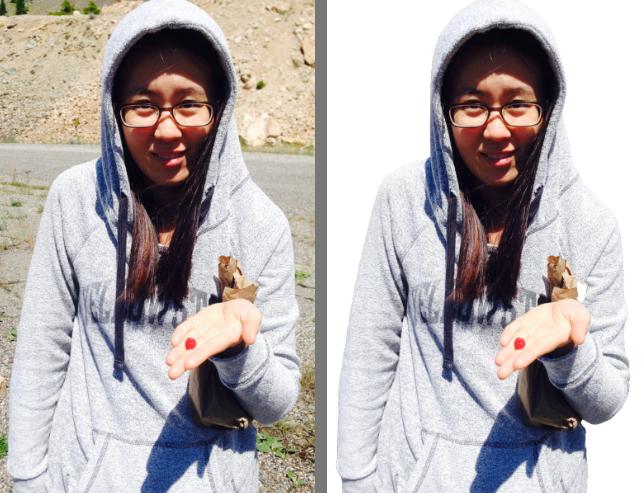} \hfill
  \includegraphics[width=0.31\linewidth, height=0.17\linewidth]{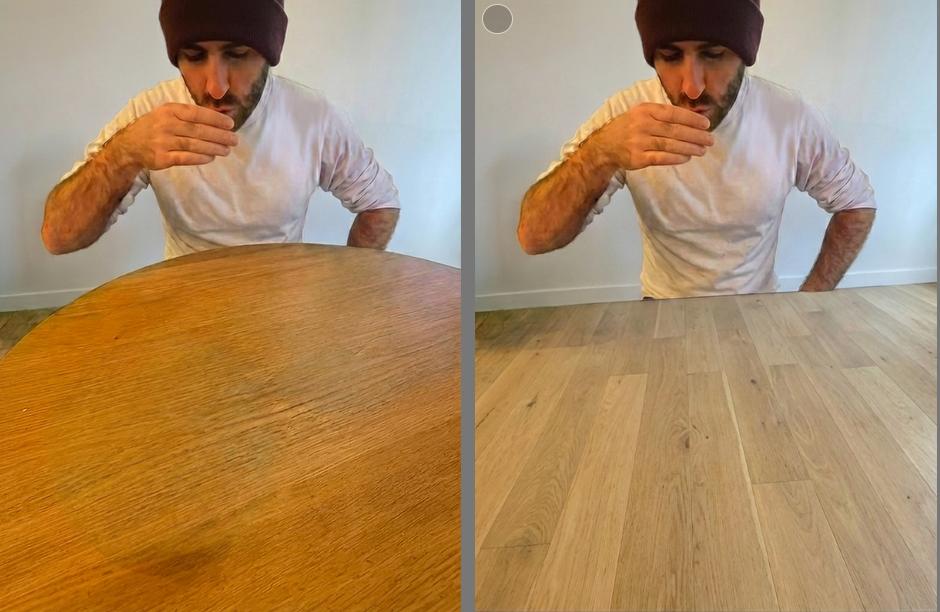} \\
    \vspace{0.02cm}
  \includegraphics[width=0.31\linewidth, height=0.17\linewidth]{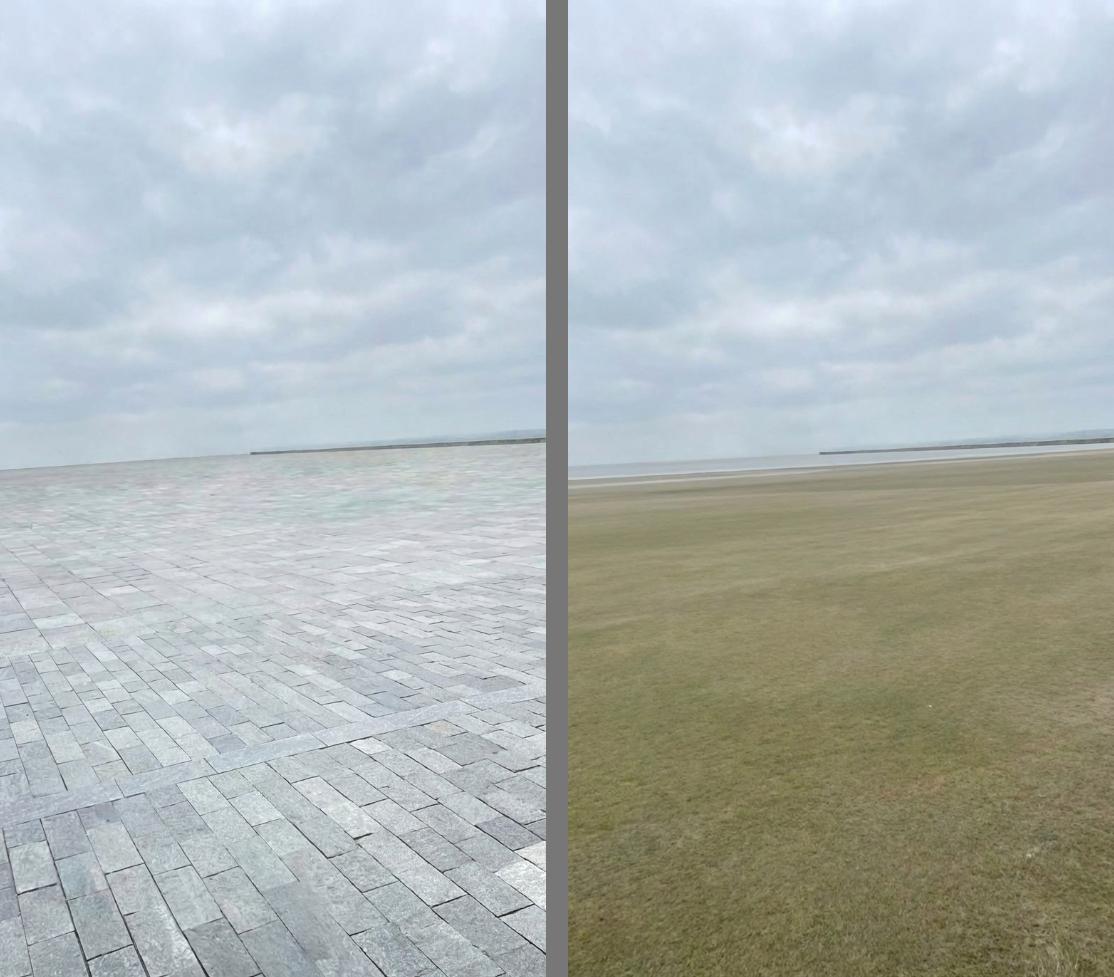} \hfill
  \includegraphics[width=0.31\linewidth, height=0.17\linewidth]{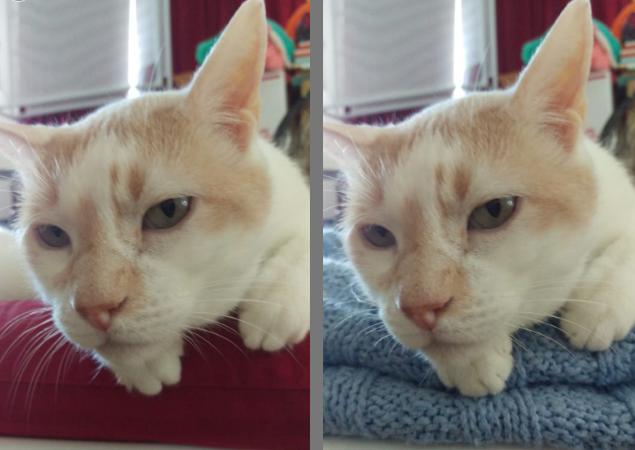} \hfill
  \includegraphics[width=0.31\linewidth, height=0.17\linewidth]{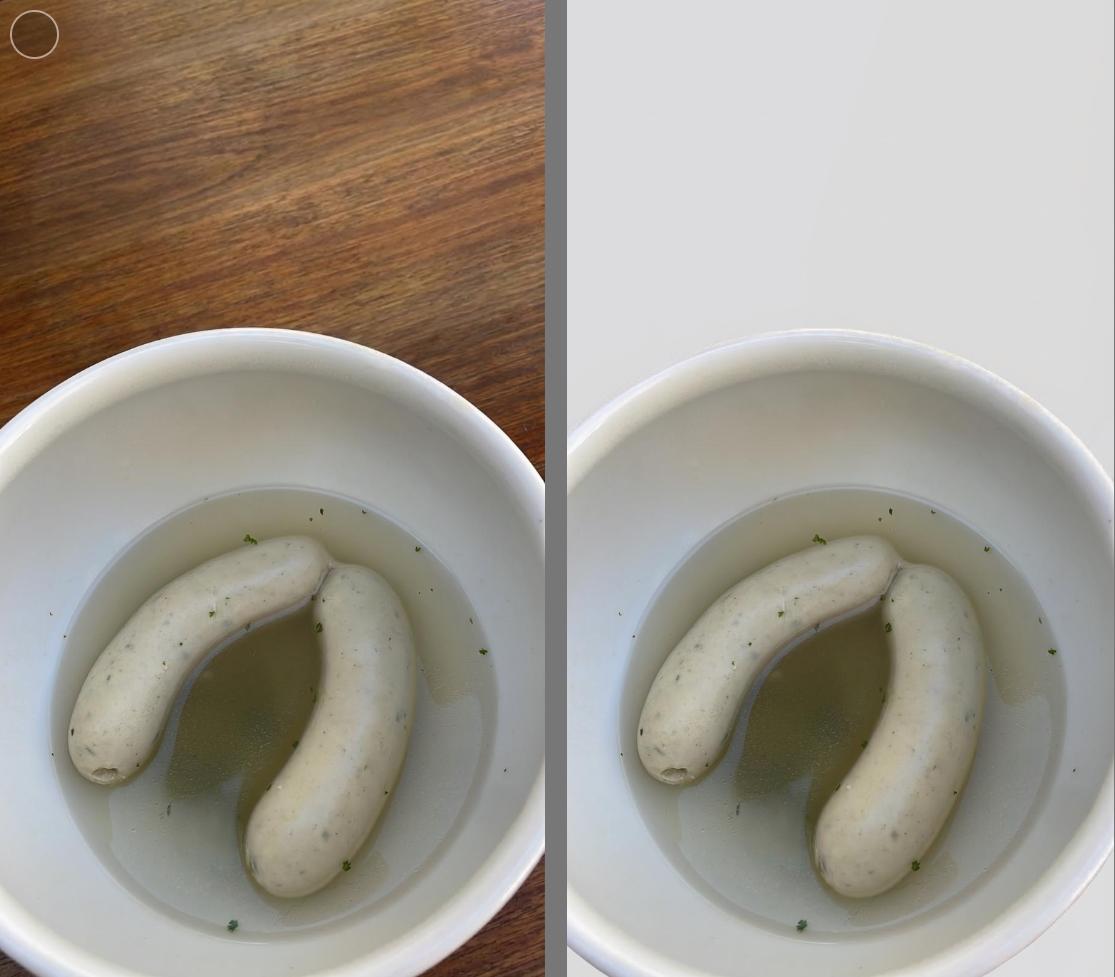} \\
  \caption{\textbf{Filtered Bad Edits.} Examples of candidate removals generated by the inpainting model that were successfully rejected by our verification classifier.}
  \label{fig:rejected_edits}
\end{figure*}

\begin{figure*}[htbp]
  \centering
    \makebox[0.16\linewidth][c]{$t_0$}%
     \makebox[0.16\linewidth][c]{$t_{10}$}%
     \makebox[0.16\linewidth][c]{$t_{15}$}%
     \makebox[0.16\linewidth][c]{$t_{20}$} %
     \makebox[0.16\linewidth][c]{$t_{25}$} %
     \makebox[0.16\linewidth][c]{$t_{30}$}\\
     \vspace{0.01cm}
  
  \includegraphics[width=\linewidth,height=0.18\linewidth]{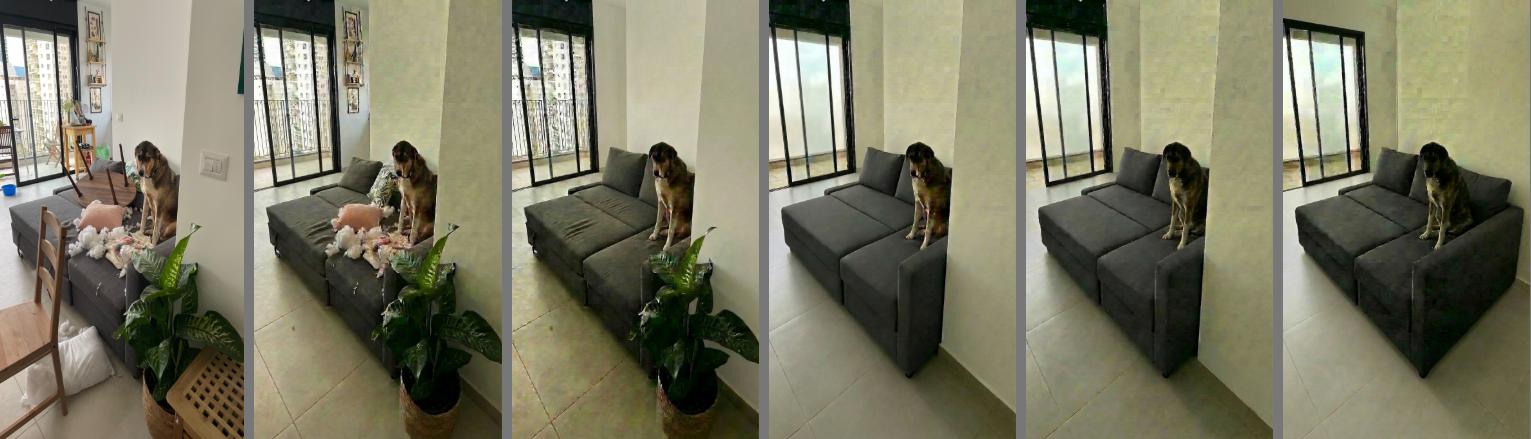} \\
  \vspace{0.02cm} 
  
  \includegraphics[width=\linewidth,height=0.18\linewidth]{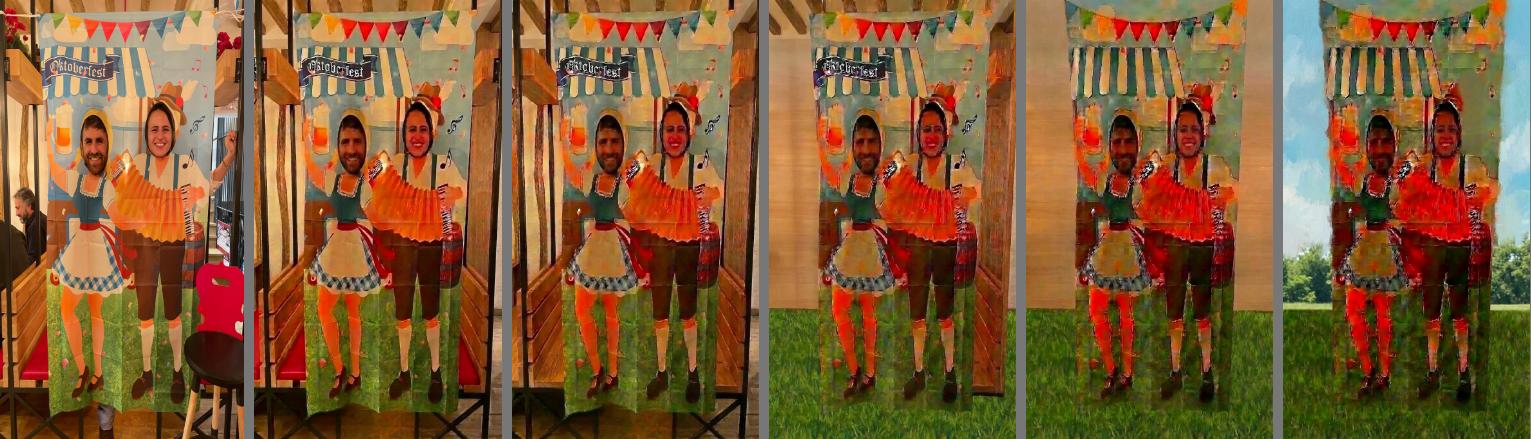} \\
  \vspace{0.02cm} 
  
  \includegraphics[width=\linewidth,height=0.18\linewidth]{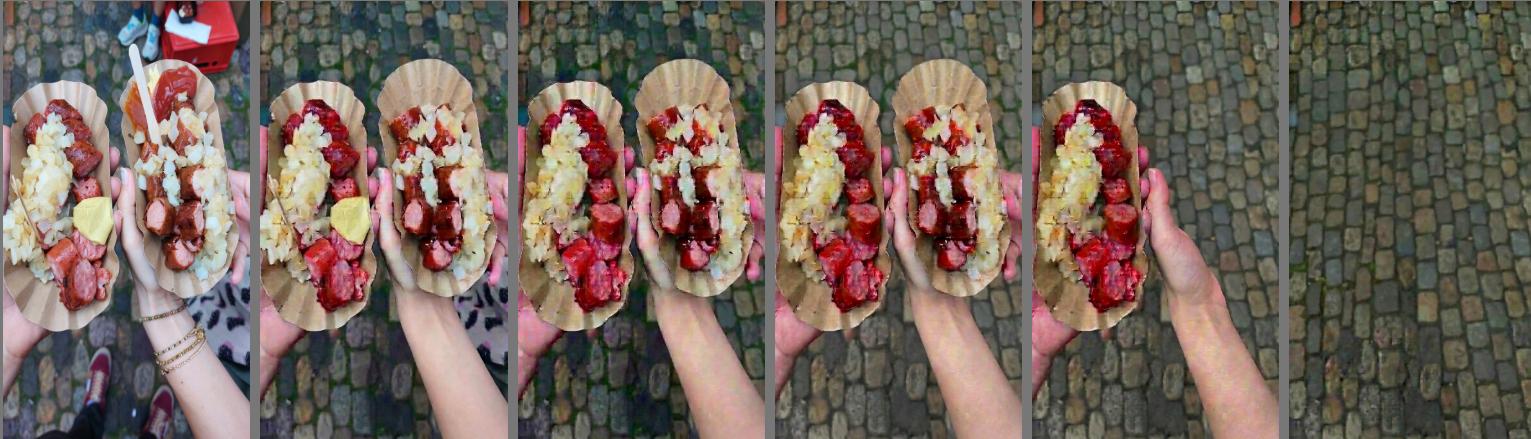} \\
  \vspace{0.02cm} 
  
  
  \includegraphics[width=\linewidth,height=0.18\linewidth]{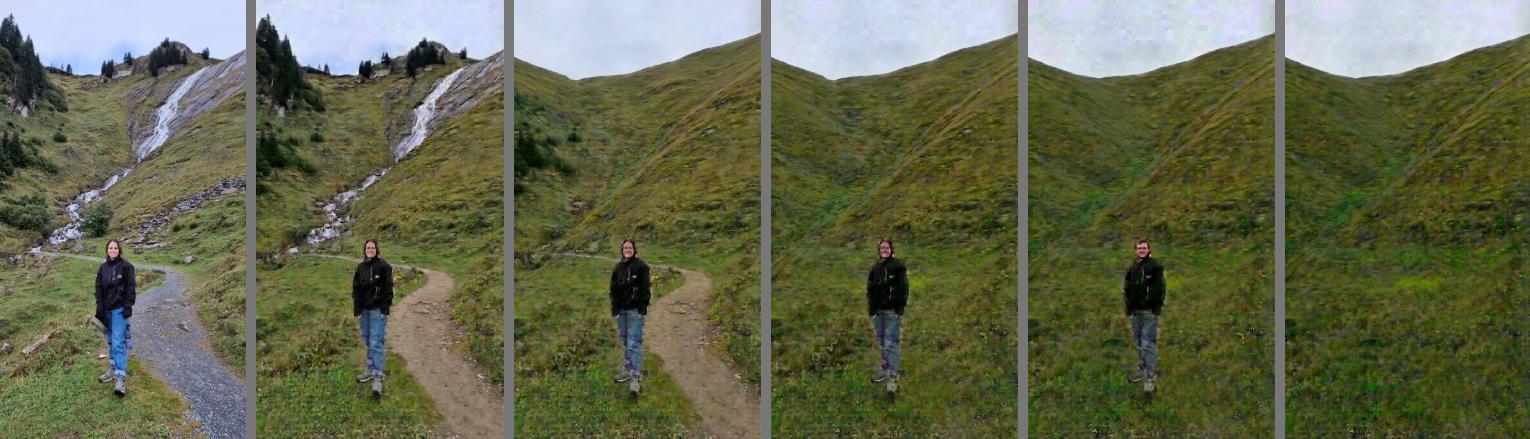} \\
  
  \caption{\textbf{Iterative Prompting Degradation.} When performing step-by-step removals using standard image editing, global image quality rapidly diverges.}
  \label{fig:iterative_degradation}
\end{figure*}

\begin{figure*}[t]
  \centering
  
    \includegraphics[width=\linewidth, height=0.2\linewidth]{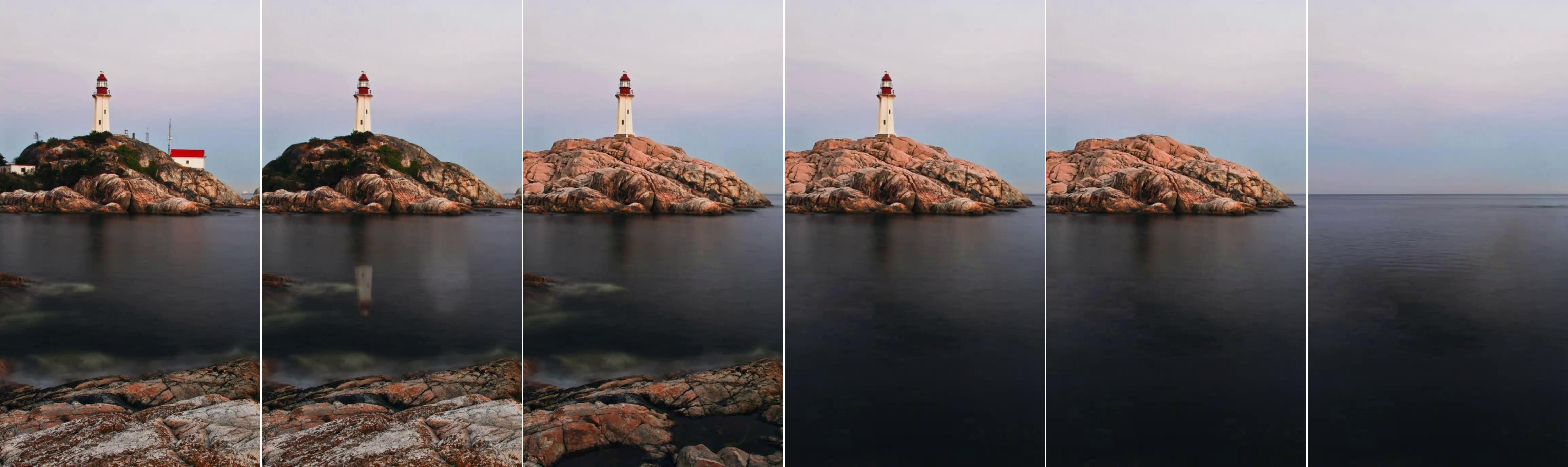}
    \vspace{0.7mm}

    \includegraphics[width=\linewidth, height=0.2\linewidth]{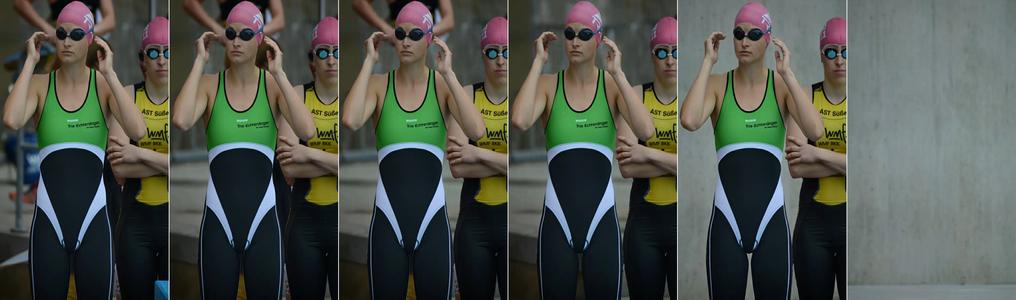}
    \vspace{0.7mm}

    \includegraphics[width=\linewidth, height=0.2\linewidth]{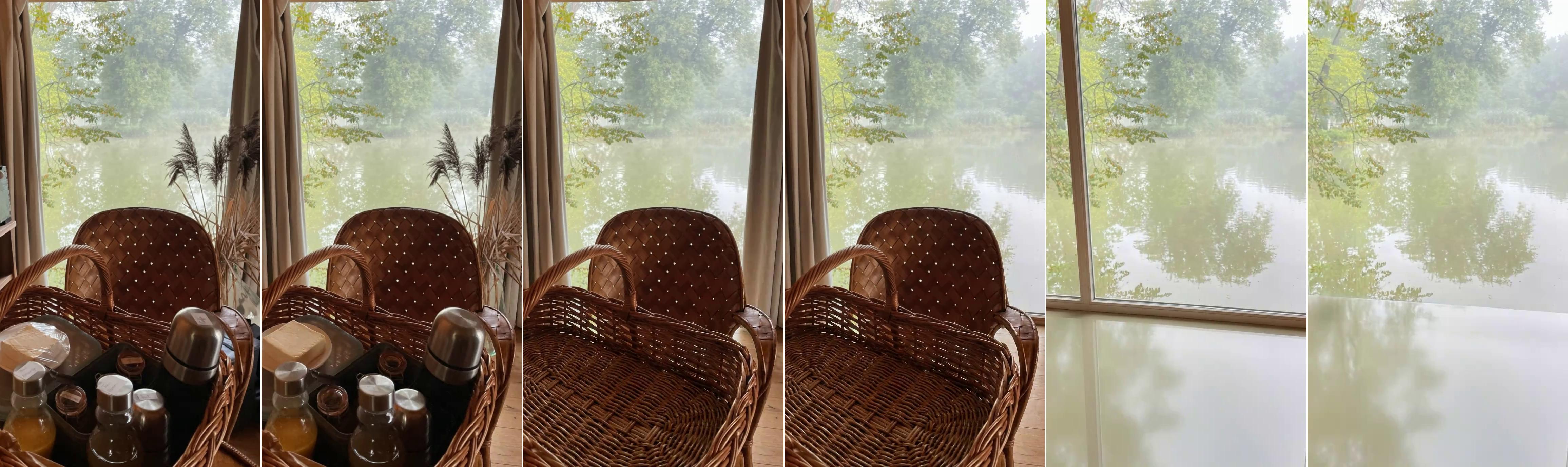}
    \vspace{0.7mm}
   
    \includegraphics[width=\linewidth]{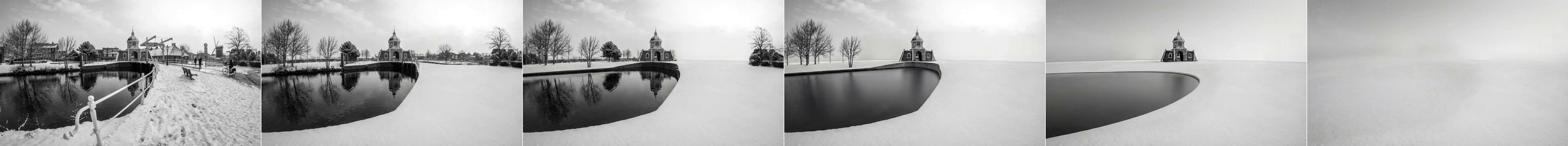}
    \vspace{0.7mm}
  
    \includegraphics[width=\linewidth]{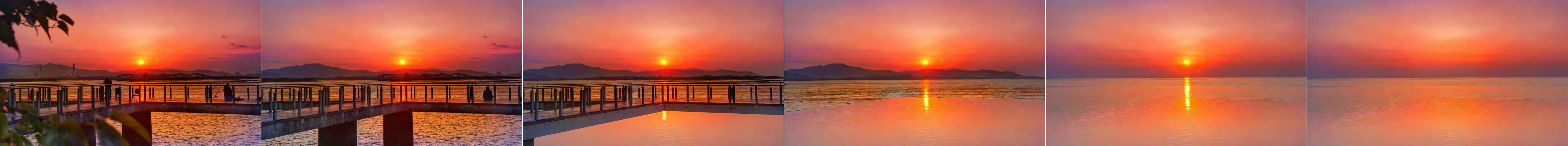}
    \vspace{0.7mm}
  
    \includegraphics[width=\linewidth]{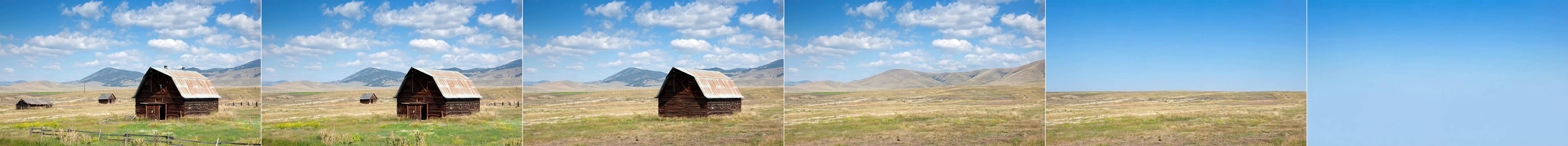}
    \vspace{0.7mm}
    
    \includegraphics[width=\linewidth]{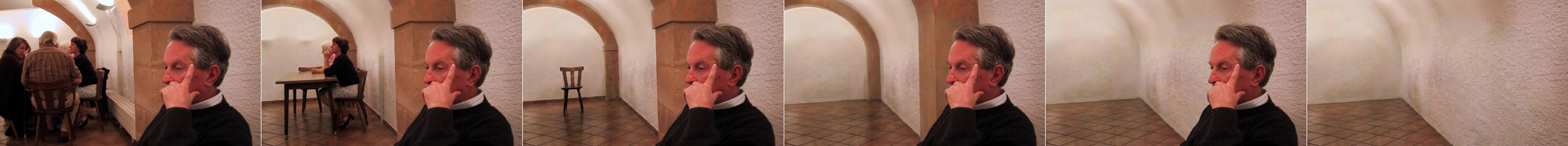}
 
    \caption{\textbf{More progressive simplification results.}}
  \label{fig:qualitative_part2}
\end{figure*}

\begin{figure*}[t]
  \centering
  
    \includegraphics[width=\linewidth, height=0.2\linewidth]{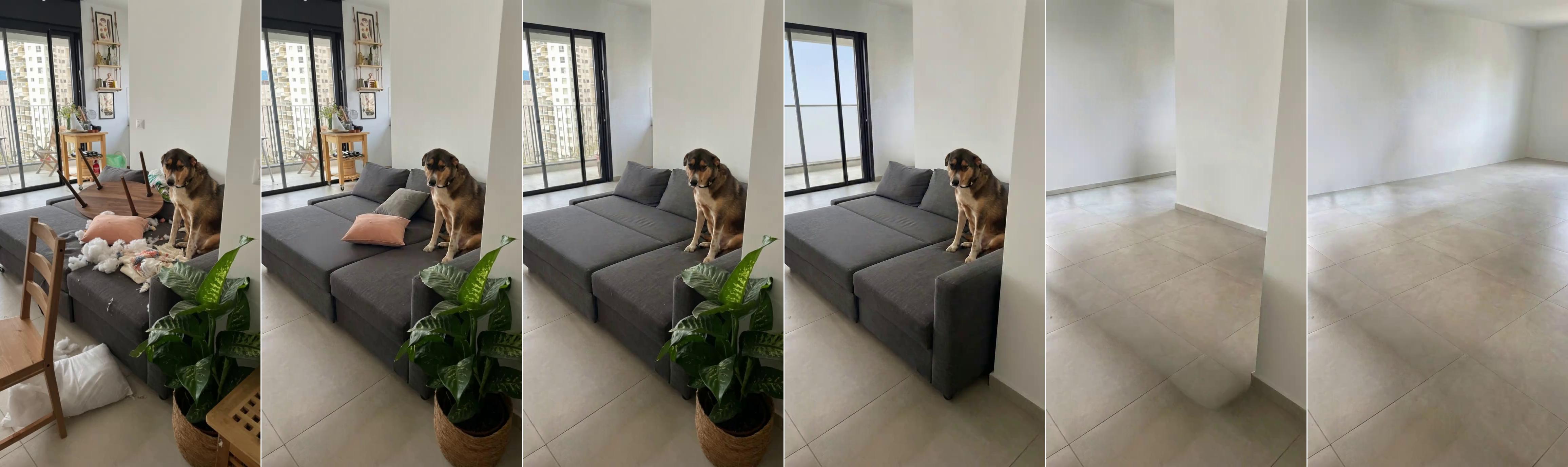}
  \vspace{0.7mm}

    \includegraphics[width=\linewidth, height=0.2\linewidth]{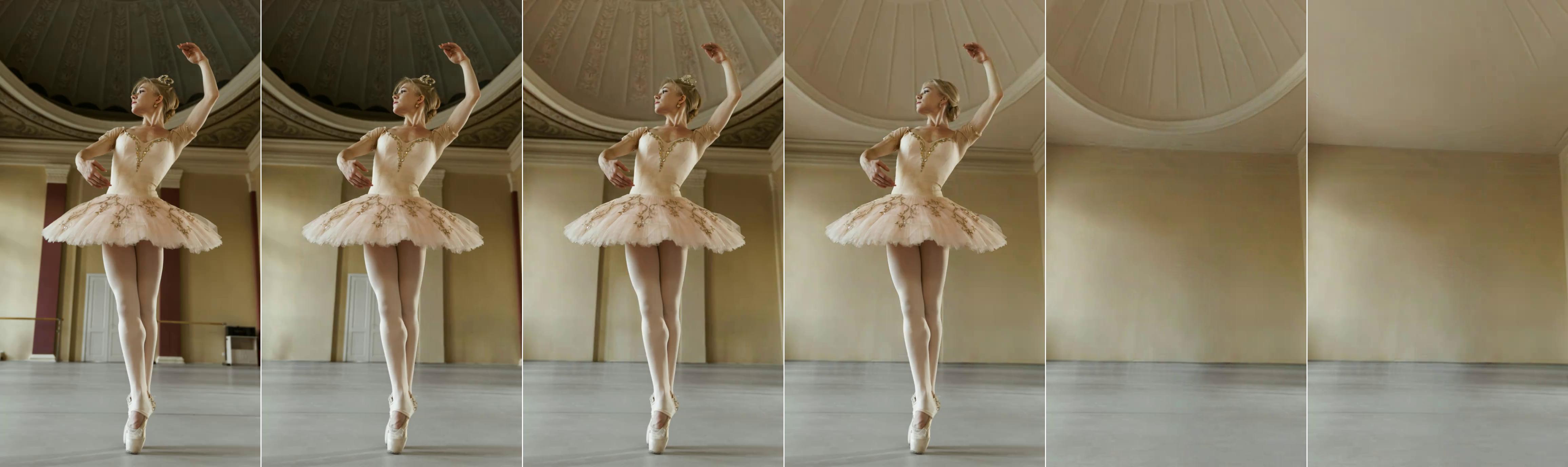}
  \vspace{0.7mm}

    \includegraphics[width=\linewidth, height=0.2\linewidth]{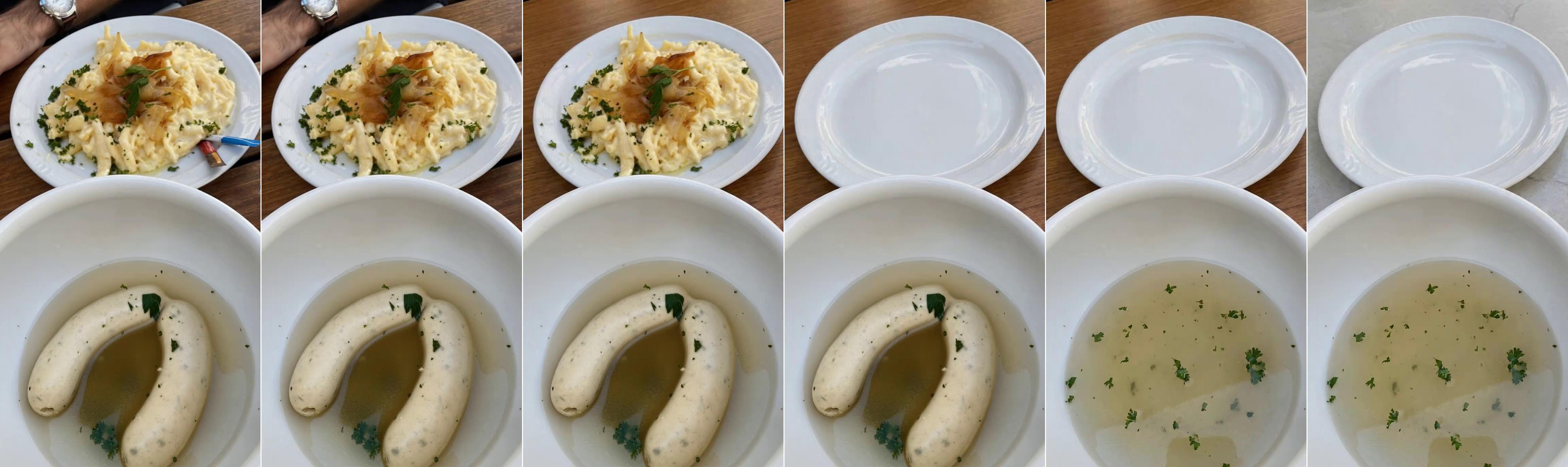}
  \vspace{0.7mm}

    \includegraphics[width=\linewidth]{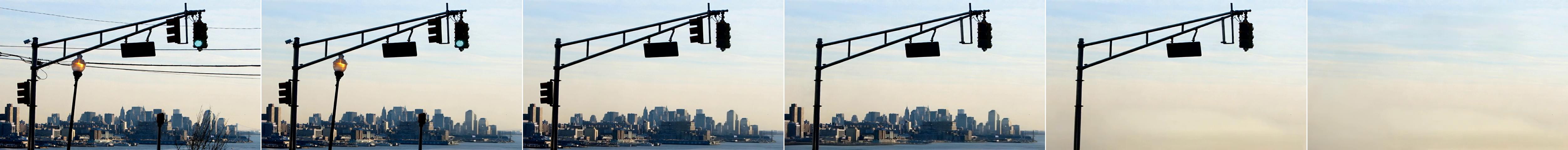}
  \vspace{0.7mm}
  
    \includegraphics[width=\linewidth]{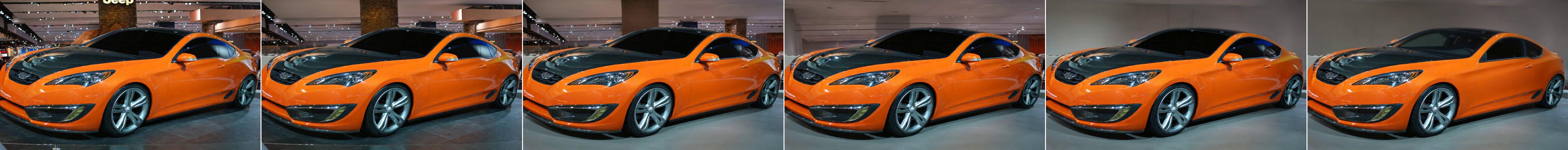}
  \vspace{0.7mm}
      
    \includegraphics[width=\linewidth]{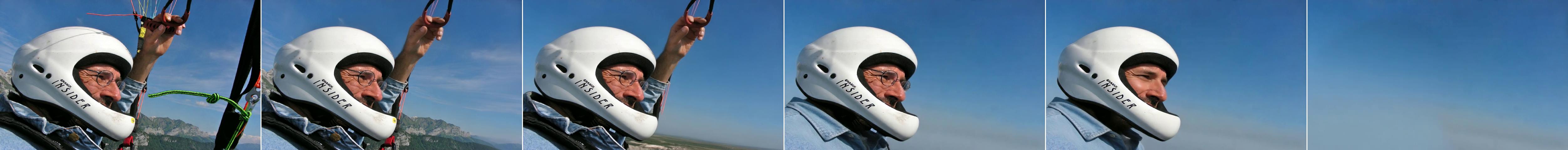}
  \vspace{0.7mm}
  
    \includegraphics[width=\linewidth]{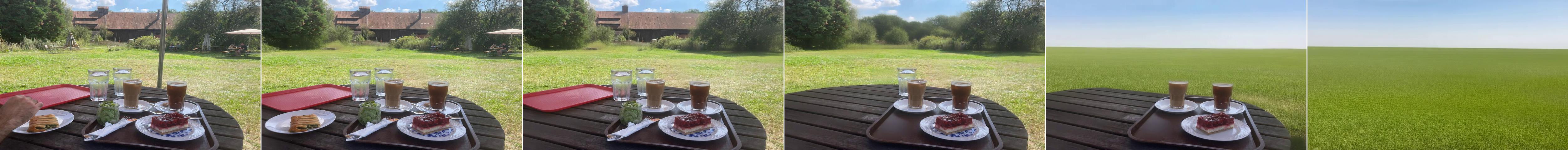}
      \caption{\textbf{More progressive simplification results.}}
  \label{fig:qualitative_part2}
\end{figure*}

\begin{figure*}[t]
  \centering
  \includegraphics[width=\linewidth, height=0.2\linewidth]{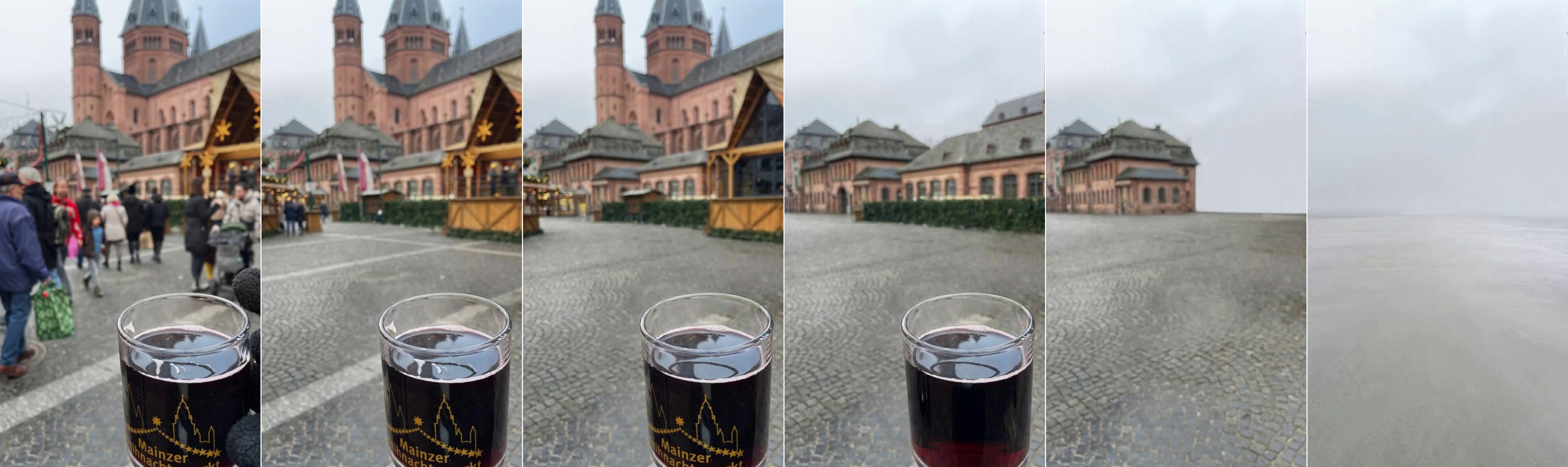}
  \vspace{0.7mm}
  
  \includegraphics[width=\linewidth, height=0.2\linewidth]{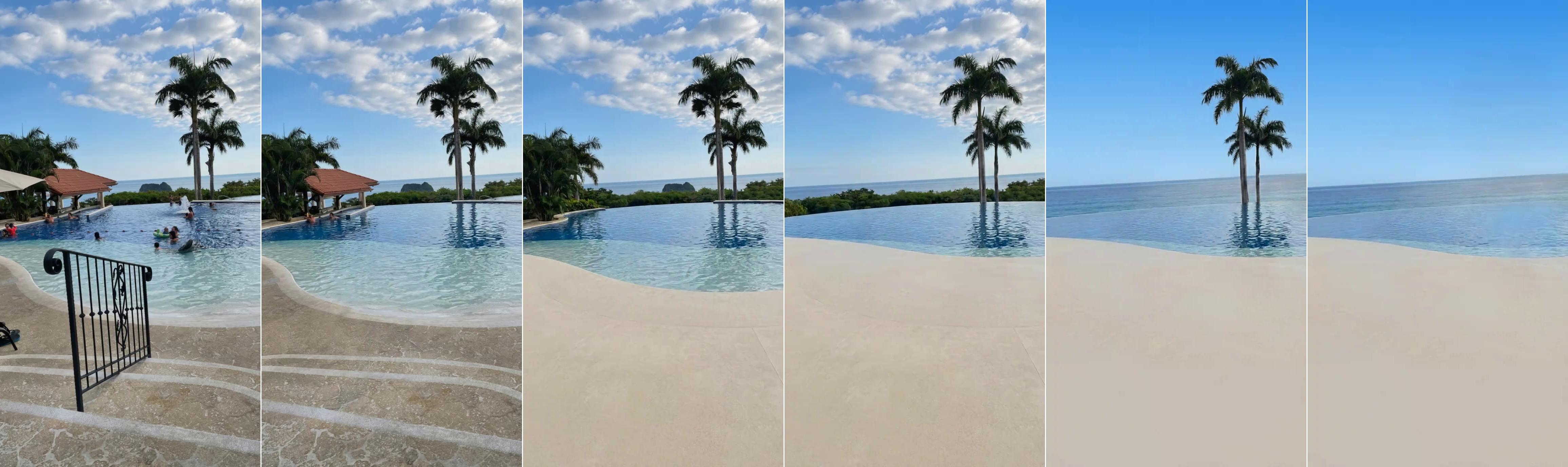}
  \vspace{0.7mm}
  
  \includegraphics[width=\linewidth, height=0.2\linewidth]{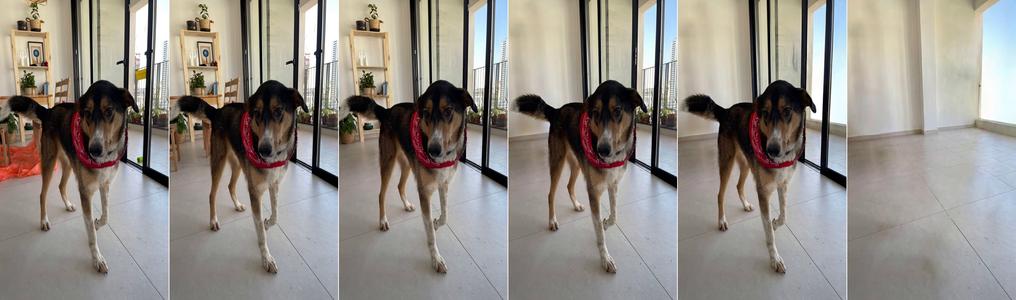}
  \vspace{0.7mm}
  
  \includegraphics[width=\linewidth]{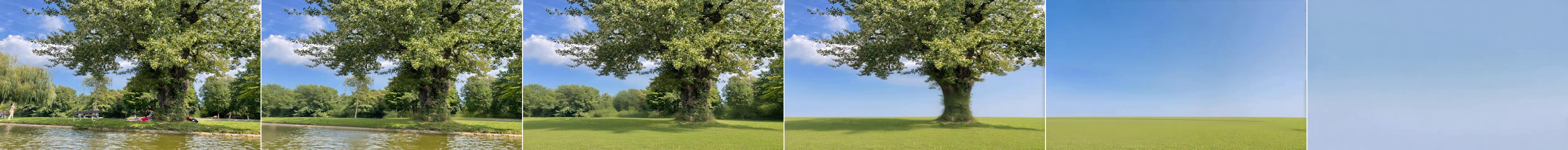}
  \vspace{0.7mm}
  
  \includegraphics[width=\linewidth]{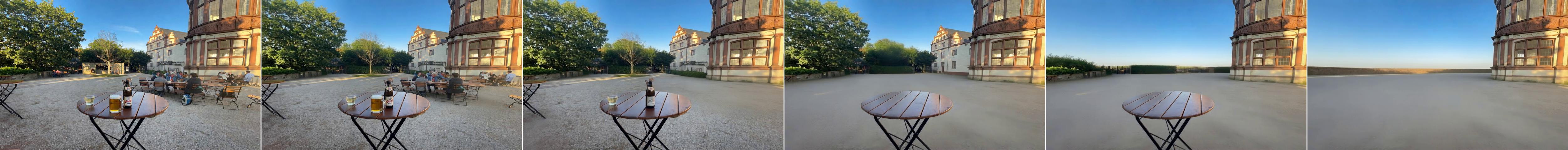}
  \vspace{0.7mm}
  
  \includegraphics[width=\linewidth]{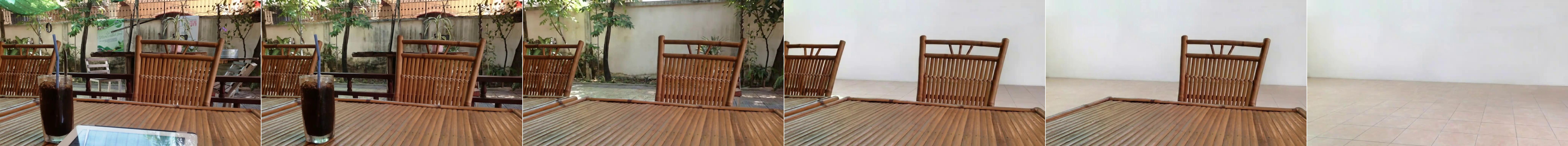}
  \vspace{0.7mm}
  
  \includegraphics[width=\linewidth]{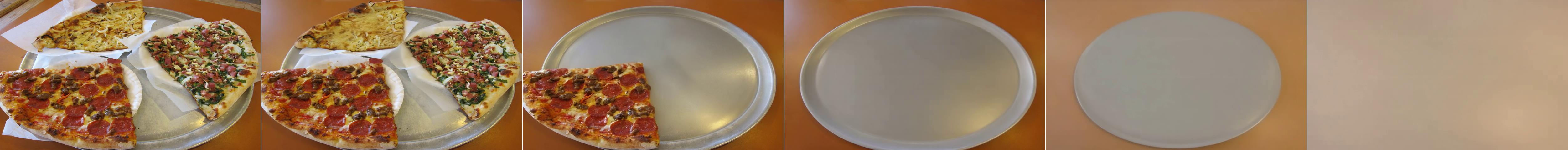}
  \caption{\textbf{More progressive simplification results.}}
  \label{fig:qualitative_part2}
\end{figure*}


\end{document}